\documentclass{article}

\PassOptionsToPackage{numbers}{natbib}

\usepackage[main, final]{neurips_2026}

\usepackage[utf8]{inputenc} 
\usepackage[T1]{fontenc}    
\usepackage{hyperref}       
\usepackage{url}            
\usepackage{booktabs}       
\usepackage{amsfonts}       
\usepackage{nicefrac}       
\usepackage{microtype}      
\usepackage{xcolor}         
\usepackage{amsmath}
\usepackage{graphicx}
\usepackage{subcaption}
\graphicspath{{figures/}}
\usepackage{multirow}
\usepackage{amsthm}
\newtheorem{proposition}{Proposition}

\newcommand*{\QED}{\null\nobreak\hfill\ensuremath{\square}}%

\title{Two-Level Softmax Sampling Done Right: Correcting~Bias~from~Size~Imbalance~and~Dispersion}

\author{%
  Walid Bendada\\
  Spotify\\
  \texttt{walidb@spotify.com} \\
   \And
   Guillaume Salha-Galvan \\
   SJTU Paris Elite Institute of Technology \\
   \texttt{gsalhagalvan@sjtu.edu.cn} \\
}

\begin{document}

\maketitle

\begin{abstract}
Sampling from a softmax distribution is a fundamental operation in machine learning, but its linear complexity in the number of items makes exact sampling impractical at scale. Two-level softmax (2LS) sampling is a popular alternative enabling sublinear-time sampling.  Assuming items are partitioned into clusters, 2LS first samples a cluster and then an item within it. In this paper, we show that, despite its advantages, 2LS introduces systematic and undesirable sampling biases, which arise from misweighting clusters by ignoring both cluster size imbalance and intra-cluster similarity dispersion. We propose two sampling methods, Size-Corrected 2LS (S-2LS) and Size- and Dispersion-Corrected 2LS (SD-2LS), which correct these biases and provide provably better softmax approximations with negligible to non-existent computational overhead. In-depth experiments on five large-scale datasets validate the improved sampling properties of our methods. We recommend their consistent use in place of standard 2LS in future work.
\end{abstract}

\section{Introduction}

Sampling items according to their similarity to a query vector in a shared embedding space is a fundamental operation in many machine learning applications, ranging from language modeling to information retrieval and recommendation ~\cite{bishop2006pattern,chen2019top,covington2016deep,vaswani2017attention}. A standard and widely used approach is to sample from a \textit{softmax distribution}, where each item is weighted by the exponential of its similarity to the query \cite{goodfellow2016deep}. However, computing and sampling from the full softmax distribution scales linearly with the number of items and quickly becomes intractable in large-scale settings \cite{bendada2025exploring,chen2016strategies,jean2015using}. As an illustration, consider a recommender system where users and items (e.g., products, videos, or music tracks) are embedded in a shared vector space, and recommendations are generated by sampling items according to their similarity to a query user embedding \cite{covington2016deep,he2017neural}. Modern recommender systems typically operate over millions of candidate items, making exact softmax sampling impractical~\cite{bendada2025exploring}.

A popular alternative to address this issue is \textit{two-level softmax} (2LS) sampling \cite{fastvae2022,midx2025,goodman2001classes,johnson2019billion,subbiah2025efficient,flashhead2026}. Assuming items are partitioned into clusters, 2LS proceeds in two steps: (1) sampling a cluster according to a softmax over cluster representatives, e.g., centroids, and (2) sampling an item within the selected cluster via a softmax over its items. As detailed in Section \ref{sec:preliminaries}, 2LS has been widely adopted across applications and offers several advantages. First and foremost, this  decomposition enables sublinear-time sampling with respect to the number of items \cite{jegou2010product,morin2005hierarchical}. Second, sampling probabilities still account for item–query similarity, as in standard softmax, unlike other scalable alternatives such as $\varepsilon$-greedy methods \cite{sutton2018reinforcement}. Third, 2LS allows sampling over the full set of items, in contrast to truncated softmax approaches that restrict sampling to a limited subset of~items~\cite{bendada2025exploring}.

In this paper, we show that, despite these advantages, 2LS exhibits important and systematic biases that, to our knowledge, have never been formally characterized. Backed by rigorous analysis, we propose corrected 2LS variants with improved properties. Specifically, our contributions~are~as~follows:
\begin{itemize}
\item We show that 2LS oversamples or undersamples items compared to softmax and, in particular, breaks its invariance: while softmax assigns identical sampling probabilities to items with equal similarity to the query, 2LS can assign substantially different probabilities to such items. This mismatch is undesirable in practice: in the recommendation example above, it implies that items equally relevant to a user may be recommended with different~probabilities.
\item We prove that 2LS systematically misweights clusters along two complementary axes. First, it ignores cluster size imbalance, oversampling small clusters and undersampling large ones relative to their true softmax mass. Second, it ignores intra-cluster similarity dispersion, undersampling clusters whose items exhibit high variability in their similarity~to~the~query.
\item We introduce Size-Corrected 2LS (S-2LS) and Size- and Dispersion-Corrected 2LS (SD-2LS), two novel sampling methods that explicitly aim to address these issues. We demonstrate that these corrections effectively mitigate the systematic biases of standard 2LS while incurring negligible to non-existent additional computational overhead.
\item We report experiments on five large-scale datasets, consistently validating the improved sampling properties of our approaches and their higher fidelity to the softmax distribution. Our code is available at: \url{https://github.com/bendadaw/neurips-2ls}.
\end{itemize}

In summary, our theoretical and empirical results show that S-2LS and SD-2LS consistently outperform 2LS in approximating softmax sampling, and we recommend their consistent use in future work. The remainder of this paper is organized as follows. Section~\ref{sec:preliminaries} introduces softmax and 2LS sampling more formally. Section~\ref{sec:bias} analyzes the biases of 2LS. Section~\ref{sec:sd2ls} introduces and studies our proposed S-2LS and SD-2LS corrections. Section~\ref{sec:experiments} presents our experiments, and Section~\ref{sec:conclusion} concludes.

\section{Preliminaries}
\label{sec:preliminaries}

\subsection{Softmax Sampling}
\label{subsec:softmax}

\paragraph{Notation}
Throughout this paper, we consider a set $\mathcal{I} = \{1, 2, \ldots, N\}$ of $N > 1$ items. Each item $i \in \mathcal{I}$ is represented by a low-dimensional vector $x_i \in \mathbb{R}^d$, i.e., an \textit{embedding},  with $d \ll N$. We do not make assumptions regarding the specific methods used to learn these embeddings. We also consider a query vector $q \in \mathbb{R}^d$ in the same embedding space. Our goal is to sample items according to their similarity to the query, as measured by the dot product $q^\top x_i$ for all $i \in \mathcal{I}$.

\paragraph{Softmax Sampling}
A standard approach is to sample from the \textit{softmax distribution}~\cite{goodfellow2016deep},~defined~as:
\begin{equation}
\forall i \in \mathcal{I},  ~p(i \mid q)
=
\frac{\exp\left(q^\top x_i / \tau\right)}
{\sum_{j=1}^N \exp\left(q^\top x_j / \tau\right)} \in [0, 1],
\label{eq:softmax}
\end{equation}
where $\tau > 0$ is a temperature parameter controlling the sharpness of the distribution. The softmax distribution assigns higher probability to items with larger dot product similarity to the query. It has become ubiquitous in machine learning, appearing in applications such as classification, language modeling, reinforcement learning, information retrieval, and recommendation~\cite{bendada2025exploring,bishop2006pattern,chen2019top,covington2016deep,sutton2018reinforcement,vaswani2017attention}.

\paragraph{Sampling Complexity}
Sampling from the softmax distribution requires computing $q^\top x_j$ scores for all $j \in \mathcal{I}$, from which both the numerators and the normalization constant are derived. Consequently, exact sampling has $\mathcal{O}(d N)$ time complexity per query~\cite{chen2016strategies,jean2015using}, which quickly becomes prohibitive in large-scale settings, such as the recommendation example with millions of items discussed in the introduction~\cite{bendada2025exploring}. This challenge has motivated approximate methods, including 2LS, presented next.

\subsection{Two-Level Softmax (2LS) Sampling}
\label{subsec:2ls}

\paragraph{Notation}
In addition to the notation of Section~\ref{subsec:softmax}, we assume that the item set $\mathcal{I}$ is partitioned into $K < N$ disjoint clusters, and denote these clusters by $\{\mathcal{C}_1, \dots, \mathcal{C}_K\}$. We have $\bigcup_{k=1}^K \mathcal{C}_k = \mathcal{I}$ and $\mathcal{C}_k \cap \mathcal{C}_{k'} = \emptyset$ for $k \neq k'$. For each cluster $\mathcal{C}_k$, we define its centroid as
$\mu_k = \frac{1}{|\mathcal{C}_k|} \sum_{i \in \mathcal{C}_k} x_i \in \mathbb{R}^d$.

\paragraph{Two-Level Softmax (2LS) Sampling} 
The objective of 2LS \cite{midx2025,goodman2001classes,johnson2019billion,subbiah2025efficient} is to first sample a cluster index $k \in \{1,\dots,K\}$, and then an item $i \in \mathcal{C}_k$ \textit{within the selected cluster}, thereby factorizing the sampling distribution as $p_{\text{2LS}}(i \mid q) = p_{\text{2LS}}(k \mid q)\, p_{\text{2LS}}(i \mid k,q)$. Clusters are sampled using a softmax over centroids, while items are sampled using another softmax restricted to the~selected~cluster:
\begin{equation}
p_{\text{2LS}}(k \mid q)
=
\frac{\exp(q^\top \mu_k / \tau)}
{\sum_{k'=1}^K \exp(q^\top \mu_{k'} / \tau)},
\text{ for } k \in \{1,\dots,K\},
\end{equation}
\begin{equation}
p_{\text{2LS}}(i \mid k,q)
=
\frac{\exp(q^\top x_i / \tau)}
{\sum_{j \in \mathcal{C}_k} \exp(q^\top x_j / \tau)},
\text{ for } i \in \mathcal{C}_k.
\end{equation}

\paragraph{Sampling Complexity}
Sampling a cluster index requires computing a softmax over $K$ centroids, with complexity $\mathcal{O}(dK)$. Sampling an item within the selected cluster $\mathcal{C}_k$ then requires computing a softmax over $|\mathcal{C}_k|$ items, with complexity $\mathcal{O}(d|\mathcal{C}_k|)$. Therefore, the cost per query of 2LS is $\mathcal{O}(dK + d|\mathcal{C}_k|)$, which is sublinear in $N$ when $K \ll N$ and clusters are sufficiently balanced. In particular, if clusters have comparable sizes, $|\mathcal{C}_k| \approx N/K$, yielding an average complexity of $\mathcal{O}(dK + dN/K)$. This is minimized when $K \approx \sqrt{N}$, leading to a complexity of $\mathcal{O}(d\sqrt{N})$.

The item set partition, required prior to sampling, can be obtained using clustering procedures such as $k$-means on embeddings~\cite{bishop2006pattern}. While this partitioning step can be costly, it is typically performed offline.
In many applications, the same partition is reused online across many sampling operations: in recommendation, items are repeatedly sampled for many queries~\cite{subbiah2025efficient}; in language modeling, tokens are repeatedly sampled for sequence generation~\cite{shao2025vq}. In such regimes, the dominant cost is thus the \textit{inference complexity for a fixed partition}, making the sublinearity derived above~a~key~advantage.

\subsection{Related Work}  
\label{subsec:relatedwork}

\paragraph{Applications of 2LS and Related Methods}
2LS and closely related variants have been widely adopted in large-scale systems where sampling over large item sets renders exact softmax computation intractable. Applications include (large) language model inference for text generation with large vocabularies~\cite{midx2025,csvdecode2025,shao2025vq,flashhead2026,dynaspec2025}, machine translation~\cite{l2s2019,dynaspec2025}, text  classification~\cite{midx2025}, image classification \cite{doublysparse2019}, and item recommendation~\cite{fastvae2022,midx2025,subbiah2025efficient}.
Despite differences in terminology and formulation, these methods share the common 2LS two-stage  structure, selecting a cluster from
a query $q$ via scores $q^\top \mu_k$ or learned variants, followed by a conditional softmax over items within~this~cluster.

2LS is also closely related to \textit{inverted file (IVF) indexing}~\cite{douze2024faiss,guo2020accelerating,jegou2010product,johnson2019billion}, an efficient approximate nearest neighbor (ANN) search technique. IVF partitions the item space into clusters, typically via $k$-means, and restricts search to a subset of clusters to retrieve similar neighbors to a query. This strategy mirrors the two-stage procedure of 2LS, where cluster selection guides subsequent item-level processing. IVF-based methods are implemented in widely used ANN libraries such as Faiss~\cite{douze2024faiss}.
2LS also shares similarities with \textit{codebook-based methods}~\cite{jegou2010product,shao2025vq}, which partition embeddings and assign items to discrete codes, while 2LS defines a probabilistic sampling procedure.

Moreover, 2LS is related to \textit{hierarchical softmax}~\cite{goodman2001classes,mnih2008scalable,mohammed2018effectiveness,morin2005hierarchical}, which factorizes the softmax along a tree, scoring items using products of conditional probabilities along root-to-leaf paths, yielding $\mathcal{O}(d\log N)$ complexity for balanced trees. In this sense, 2LS can be viewed as a two-level instance of this idea. However, while hierarchical softmax relies on predefined trees and provides an exact factorization, 2LS uses a cluster-based approximation and a query-dependent sampling~procedure.

\paragraph{Other Softmax Approximations}
For completeness, we note that a broad body of work, less directly related to 2LS, has also explored alternative methods to accelerate, approximate, or replace softmax computation for various applications \cite{chen2016strategies}. In particular, during training, approaches such as \textit{sampled softmax}~\cite{jean2015using} approximate the partition function using a subset of classes, while \textit{adaptive softmax}~\cite{joulin2017efficient} partitions classes by frequency to reduce computation in skewed (e.g., Zipfian) distributions.

In large-scale inference settings, such as large language models or recommender systems with large item sets, simple \textit{truncation-based methods} remain among the most widely used in practice. They restrict the softmax computation to the top-$k$ items in terms of similarity~\cite{bendada2025exploring,chen2021user}, with $k \ll N$, or to the smallest set of items whose cumulative probability exceeds a threshold $p < 1$, yielding nucleus (top-$p$) sampling~\cite{brown2020language,holtzman2020curious}. These methods can be combined with ANN search to reduce computational cost~\cite{chen2019top}. However, they restrict sampling to a truncated subset of items. This limitation is detrimental in applications such as recommendation, where constraining the candidate item set based on computational considerations only may exclude relevant items beyond the selected subset~\cite{bendada2025exploring}.

Finally, one may also replace softmax with  methods such as von Mises--Fisher sampling (which samples proportionally to $q^\top x_j$, in sublinear time, but requires hyperspherical assumptions \cite{bendada2024vmf,bendada2025exploring,bendada2025glove}) and $\varepsilon$-greedy sampling (which runs in constant time but ignores item-query similarity~\cite{sutton2018reinforcement}). In this context, 2LS appears as an  alternative to simultaneously (1) account for $q^\top x_j$ similarities as in softmax, (2) avoid restricting sampling to an item subset, and (3) enable sublinear-time complexity.

\section{Bias Characterization of Two-Level Softmax (2LS) Sampling}
\label{sec:bias}

In this section, we show that, despite its advantages, 2LS exhibits systematic sampling biases, and we analyze their underlying causes. 
We report all mathematical proofs in Appendix \ref{appendix:proofs-section3} of this paper.

\subsection{Sampling Ratio}
To analyze the discrepancy between exact softmax and 2LS in this paper, we introduce the ratio
\begin{equation}
R_i^{(N)}(q) = \frac{p_{\text{2LS}}(i \mid q)}{p(i \mid q)}.
\label{eq:ratio}
\end{equation}
For any item $i \in \mathcal{I}$, query $q \in \mathbb{R}^d$, and number of items $N$, this ratio captures the relative sampling distortion induced by the 2LS procedure: values greater than $1$ indicate that $i$ is oversampled by 2LS compared to the exact softmax, while values smaller than $1$ indicate undersampling.

A notable property of this ratio is that it depends only on the 2LS cluster containing $i$, and not directly on the similarity $q^\top x_i$, as formalized in Proposition~\ref{prop:cluster}. Consequently, although items within the same cluster  may have different sampling probabilities, they are over- or undersampled by the same factor.
\begin{proposition}
\label{prop:cluster}
Let $q \in \mathbb{R}^d$, and let $i, j \in \mathcal{I}$ with $i \neq j$. If $i, j \in \mathcal{C}_k$ for any $\mathcal{C}_k \in \{\mathcal{C}_1, \dots, \mathcal{C}_K\}$, i.e., if the items $i$ and $j$ belong to the same 2LS cluster, then $R_i^{(N)}(q) = R_j^{(N)}(q)$. \end{proposition} 

\subsection{Systematic Bias in 2LS Sampling}

\paragraph{Setting}
In the following, we study $R_i^{(N)}(q)$ under a general probabilistic setting where, for each $N$, item embeddings $x_1, \dots, x_N$ are i.i.d.\ samples from a mixture distribution $\sum_{k=1}^K \pi_k \mathcal{D}_k$, where $\pi_k$ denotes the probability of sampling from the component distribution $\mathcal{D}_k$, and such that, for each $k \in \{1,\dots,K\}$, the item embeddings $\{x_i : i \in \mathcal{C}_k\}$ are samples from $\mathcal{D}_k$.

\paragraph{Asymptotic Behavior} $R_i^{(N)}(q)$ is a random variable. We study its behavior as $N$ increases. Next, expectations are taken with respect to the mixture $x \sim \sum_{k=1}^K \pi_k \mathcal{D}_k$, unless specified~otherwise.

\begin{proposition}
\label{prop:main}
Let $i \in \mathcal{I}$, $q \in \mathbb{R}^d$, and $k$ be the cluster index such that $x_i \in \mathcal C_{k}$. Assume $\pi_k>0$.~Then,
\begin{equation}
R_i^{(N)}(q)
\xrightarrow[N \to \infty]{a.s.} R_k^{(\infty)}(q)
=
c(q)\,
\frac{\exp(q^\top \mu_k / \tau)}
{\pi_k\,\mathbb E_{x\sim \mathcal D_k}[\exp(q^\top x / \tau)]}, \text{ with }
c(q)
=
\frac{\mathbb E[\exp(q^\top x / \tau)]}
{\sum\limits_{k'=1}^K \exp(q^\top \mu_{k'} / \tau)}.
\end{equation}
\end{proposition}

\begin{proposition}[Corollary with Gaussian approximation]
\label{corollary:gaussian}
In addition, assume that $q^\top x_i/\tau$ follows a Gaussian distribution with mean $q^\top \mu_{k}/\tau$ and variance $\sigma_k^2(q) = \mathrm{Var}_{x\sim \mathcal D_k}(q^\top x/\tau)$. Then,
\begin{equation}
R_k^{(\infty)}(q)
\propto
\frac{1}{\pi_k\,\exp\!\left(\tfrac{1}{2}\sigma_k^2(q)\right)}.
\end{equation}
\end{proposition}

\paragraph{Discussion}
Our results show that $R_k^{(\infty)}(q) \neq 1$ in general, i.e., that 2LS systematically oversamples or undersamples items compared to exact softmax sampling. In particular, two items $i, j \in \mathcal{I}$ with equal similarity to the query $q$, i.e., $q^\top x_i = q^\top x_j$, would receive identical sampling probabilities under softmax, but this invariance can be broken under 2LS if $i$ and $j$ belong to different clusters, as they may be oversampled or undersampled by different relative factors.

Importantly, our results provide a theoretical understanding of this phenomenon, which is particularly clear in the simplified ratio form given by our corollary under a Gaussian approximation. They show that 2LS misweights clusters along two complementary axes. First, it ignores \textit{cluster size imbalance}, thereby oversampling smaller clusters and undersampling larger ones. Second, it ignores the \textit{intra-cluster item-query similarity dispersion}, undersampling clusters whose items exhibit higher variability in their similarity to the query. Our experiments in Section~\ref{sec:experiments} will empirically confirm and illustrate the practical significance of these biases on various large-scale datasets.

To our knowledge, these biases have never been formally characterized in previous work. The closest connection is a recent observation by \citet{flashhead2026}, who empirically noted a loss of efficiency under cluster size imbalance, consistent with our findings. Our results suggest corrections accounting for cluster
size imbalance and intra-cluster dispersion, which we develop in the next section.



\section{Size- and Dispersion-Corrected Two-Level Softmax Sampling}
\label{sec:sd2ls}

We now propose two corrected sampling methods designed to address the sampling biases of 2LS. We report all mathematical proofs from this section in Appendix \ref{appendix:proofs-section4} of this paper.

\subsection{Size-Corrected Two-Level Softmax (S-2LS) Sampling}

\paragraph{S-2LS Sampling}
We first propose S-2LS, a corrected 2LS procedure that incorporates cluster size into the cluster-level score, while the within-cluster distribution remains unchanged:
\begin{equation}
\label{eq:s2ls-1}
p_{\text{S-2LS}}(k \mid q)
=
\frac{|\mathcal C_k| \exp(q^\top \mu_k/\tau)}
{\sum_{k'=1}^K |\mathcal C_{k'}| \exp(q^\top \mu_{k'}/\tau)}, \text{ for } k \in \{1,\dots,K\},
\end{equation}
\begin{equation}
p_{\text{S-2LS}}(i \mid k, q)
=
\frac{\exp(q^\top x_i/\tau)}
{\sum_{j \in \mathcal C_k} \exp(q^\top x_j/\tau)}, \text{ for } i \in \mathcal C_k.
\end{equation}

Equation \eqref{eq:s2ls-1} requires computing a softmax over $K$ $d$-dimensional dot products, with complexity $\mathcal{O}(dK)$. The within-cluster step is unchanged, so the overall sampling complexity of S-2LS is $\mathcal{O}(dK + d|\mathcal C_k|)$, i.e., identical to that of 2LS. In particular, following the reasoning of Section~\ref{subsec:2ls}, choosing $K \approx \sqrt{N}$ and balancing clusters yields an $\mathcal{O}(d \sqrt{N})$ complexity.

\paragraph{Asymptotic Behavior} Analogous to the ratio of Equation~\eqref{eq:ratio}, we introduce
$R_{i, \text{S-2LS}}^{(N)}(q) = \frac{p_{\text{S-2LS}}(i \mid q)}{p(i \mid q)}$, a variant of the same ratio for S-2LS, and study its asymptotic behavior in the setting of Section~\ref{sec:bias}.

\begin{proposition}
\label{prop:first_order_ratio}
Let $i \in \mathcal{I}$, $q \in \mathbb{R}^d$, and $k$ be such that $x_i \in \mathcal C_{k}$. Assume $\pi_k > 0$. Then,
\begin{equation}
R_{i, \text{S-2LS}}^{(N)}(q)
\xrightarrow[N \to \infty]{a.s.}
c_{\text{S-2LS}}(q)\,
\frac{\exp(q^\top \mu_k/\tau)}
{\mathbb E_{x\sim \mathcal D_k}[\exp(q^\top x/\tau)]},
\end{equation}
where $c_{\text{S-2LS}}(q)$ does not depend on $i$ or $k$.
\end{proposition}

In summary, the proposed S-2LS correction removes the dependence on cluster size. Our experiments will confirm this result empirically across several datasets, showing that S-2LS does not undersample (respectively, oversample) items from larger (resp., smaller) clusters.

\subsection{Size- and Dispersion-Corrected Two-Level Softmax (SD-2LS) Sampling}

\paragraph{SD-2LS Sampling} S-2LS alone does not address the second source of bias identified in Section~\ref{sec:bias}, namely the variability of intra-cluster item-query similarities. We therefore propose a second method, SD-2LS, which, in addition to accounting for cluster size imbalance as in S-2LS, also accounts for dispersion. Denoting $\sigma_k^2(q) = \mathrm{Var}_{x\sim \mathcal D_k}(q^\top x/\tau)$, we propose to sample items as follows:
\begin{equation}
p_{\text{SD-2LS}}(k \mid q)
=
\frac{|\mathcal C_k| \exp\!\left(q^\top \mu_k/\tau + \tfrac12 \sigma_k^2(q)\right)}
{\sum_{k'=1}^K |\mathcal C_{k'}| \exp\!\left(q^\top \mu_{k'}/\tau + \tfrac12 \sigma_{k'}^2(q)\right)} \text{ for } k \in \{1,\dots,K\},
\end{equation}
\begin{equation}
p_{\text{SD-2LS}}(i \mid k, q)
=
\frac{\exp(q^\top x_i/\tau)}
{\sum_{j \in \mathcal C_k} \exp(q^\top x_j/\tau)}, \text{ for } i \in \mathcal C_k.
\end{equation}
 In addition to dot products $q^\top \mu_k$, this variant requires evaluating the quadratic forms $q^\top \Sigma_k q$ for all $K$ clusters, where the $d \times d$ matrices $\Sigma_k$ denote the intra-cluster covariance matrices, yielding an $\mathcal{O}(d^2K)$ cost. The within-cluster step is unchanged, so the overall sampling cost is $\mathcal{O}(d^2 K+ d |\mathcal C_k|)$. In practice, since $d$ is typically much smaller than $N$, which can reach millions, the additional cost compared to 2LS remains negligible. In particular, following the reasoning of Section~\ref{subsec:2ls}, choosing $K \approx \sqrt{N}$ and balancing clusters yields an $\mathcal{O}(d^2 \sqrt{N})$ complexity, preserving sublinear scaling in $N$.
 
\paragraph{Asymptotic Behavior} Analogous to Equation~\eqref{eq:ratio}, we introduce
$R_{i,\text{SD-2LS}}^{(N)}(q) = \frac{p_{\text{SD-2LS}}(i \mid q)}{p(i \mid q)}$, a variant of the same ratio for SD-2LS, and study its asymptotic behavior in the setting of Section~\ref{sec:bias}.

\begin{proposition}
\label{prop:second_order_ratio}
Let $i \in \mathcal{I}$, $q \in \mathbb{R}^d$, and $k$ be such that $x_i \in \mathcal C_{k}$. Assume $\pi_k > 0$. Then,
\begin{equation}
R_{i, \text{SD-2LS}}^{(N)}(q)
\xrightarrow[N \to \infty]{a.s.}
c_{\text{SD-2LS}}(q)\,
\frac{\exp\!\left(q^\top \mu_k/\tau + \tfrac12 \sigma_k^2(q)\right)}
{\mathbb E_{x\sim \mathcal D_k}[\exp(q^\top x/\tau)]},
\end{equation}
where $c_{\text{SD-2LS}}(q)$ does not depend  on $i$ or $k$.
\end{proposition}

Proposition~\ref{prop:second_order_ratio} shows that, regardless of the underlying distribution, the denominator term involves the \textit{moment generating function} (MGF) of the scalar projection $q^\top x / \tau$. This perspective provides a natural interpretation of the remaining distortion. In general, the MGF can be expanded in terms of the moments of the distribution, and is therefore well approximated by its leading terms when higher-order moments (e.g., skewness and kurtosis) are small~\cite{casella2024statistical}.

This observation suggests a natural correction strategy: ideally, one would reweight each cluster by $\mathbb{E}_{x \sim \mathcal D_k}[\exp(q^\top x / \tau)]$ to fully account for dispersion. However, computing this quantity exactly would require evaluating all similarities within each cluster, leading to a cost of $\mathcal{O}(N)$, which defeats the purpose of sublinear sampling.
Instead, in this paper, our strategy is to approximate the MGF using a \textit{truncated expansion}. In particular, retaining terms up to second order as done in SD-2LS yields a correction that depends on the variance $\sigma_k^2(q)$, which can be computed efficiently.

Notably, as formalized in Proposition~\ref{prop:second_order_ratio_gaussian}, this approximation is exact under Gaussian assumptions, as all higher-order moments beyond the variance vanish. Therefore, SD-2LS asymptotically recovers exact softmax sampling in this case. Empirically, our results on real-world datasets in the next section consistently show that this second-order correction remains effective even when the Gaussian assumption does not strictly hold, confirming the practical utility of SD-2LS on real-world data.

\begin{proposition}[Corollary with Gaussian approximation]
\label{prop:second_order_ratio_gaussian}
In addition, assume that $q^\top x_i/\tau$ follows a Gaussian distribution with mean $q^\top \mu_{k}/\tau$ and variance $\sigma_k^2(q) = \mathrm{Var}_{x\sim \mathcal D_k}(q^\top x/\tau)$.~Then, 
\begin{equation}
\forall  i \in \mathcal{I}, R_{i, \text{SD-2LS}}^{(N)}(q)
\xrightarrow[N \to \infty]{a.s.} 1.
\end{equation}
\end{proposition}

\subsection{Discussion}
\label{subsec:discussionandlimitations}
\paragraph{Usage of S-2LS vs SD-2LS} 
SD-2LS provides stronger guarantees than S-2LS; it is therefore natural to ask why S-2LS was introduced. The two variants are in fact suited to different regimes. S-2LS compensates for cluster size imbalance at the same sampling complexity as 2LS. It is the preferred choice when clusters are highly imbalanced and low-latency sampling is critical.
SD-2LS additionally accounts for intra-cluster dispersion along the query direction. This yields a more faithful softmax approximation, at the cost of slightly higher computation due to covariance evaluations, introducing a $d^2$ term in the complexity instead of $d$. While negligible in many cases, this overhead may become more significant in high-dimensional settings with large $d$, making SD-2LS preferable only when intra-cluster dispersion is substantial and sufficient resources are available.
In summary, S-2LS provides a lightweight correction, while SD-2LS trades increased computation for improved~sampling~accuracy.

\paragraph{Limitations and Future Work}

Our analysis of SD-2LS relies on truncated expansions of the MGF of the dot product similarity. While this paper shows this approximation is exact under Gaussian assumptions and empirically effective more broadly, it may be less accurate for embedding distributions exhibiting significant higher-order moments (e.g., high skewness or kurtosis). In such cases, higher-order corrections to 2LS could further improve sampling accuracy. Extending our analysis to account for these effects, while maintaining computational efficiency, is an interesting direction for future work.
Moreover, while our analysis focuses on 2LS, the bias study extends naturally to hierarchical softmax constructions based on trees~\cite{goodman2001classes,morin2005hierarchical}. In such models, the mismatch identified in Section~\ref{sec:bias} recurs at each level of the root-to-leaf hierarchy, accumulating multiplicatively along the path. This suggests that our approach could be extended to hierarchical softmax by incorporating corrections at each node, which we leave for future work.

\section{Experimental Analysis}
\label{sec:experiments}

In this section, we empirically validate our theoretical results and the practical effectiveness of the proposed methods. Our experiments are designed to address the following questions:
\begin{itemize}
\item \textbf{Q1}: Are the sampling biases of standard 2LS, theoretically characterized in Section~\ref{sec:bias}, observable and practically significant?
\item \textbf{Q2}: Do S-2LS and SD-2LS effectively correct these biases, yielding samples that more closely match the exact softmax distribution than 2LS?
\item \textbf{Q3}: How do S-2LS and SD-2LS compare in terms of the fidelity-latency trade-off relative to other approximation methods, including 2LS, hierarchical softmax, and top-$k$ softmax?
\end{itemize}

\subsection{Experimental Setup}
\label{subsec:setup}

\paragraph{Datasets} We consider five large-scale datasets, including three real-world embedding corpora and two controlled synthetic corpora. We use two item embedding datasets, \textit{VK-LSVD}~\cite{vk2026} ($N=19.6$M  videos, $d=64$) and \textit{YAMBDA}~\cite{ploshkin2025yambda} ($N=7.7$M music tracks, $d=128$), extracted from industrial recommender systems. We additionally use \textit{GloVe-100}~\cite{pennington2014glove} ($N=1.2$M word embeddings, $d=100$) to extend the evaluation to NLP data.
Finally, we construct two synthetic Gaussian mixture corpora, \textit{Synth-balanced} and \textit{Synth-unbalanced} (both $N=10^6$, $d=100$), to isolate the role of cluster-size imbalance. They are obtained by generating $1024$ mean vectors uniformly on the $d$-dimensional unit hypersphere, and forming embedding clusters by adding isotropic Gaussian noise ($\sigma=0.1$) around these means. Synth-balanced has an equal number of points per component. In contrast, Synth-unbalanced follows a heavy-tailed distribution resembling real-world corpora, with component sizes drawn from a multinomial distribution with log-normal weights $w_c \propto \exp(0.8\,Z_c)$, with $Z_c \sim \mathcal{N}(0,1)$. GloVe-100 and YAMBDA embeddings are $\ell_2$-normalized, while VK-LSVD and the synthetic corpora are left unnormalized, ensuring coverage of a broad range of regimes. 

\paragraph{Methods} We compare six sampling methods. \textit{Exact softmax} performs the full $\mathcal{O}(N)$ pass and serves as the reference. We compare it against \textit{2LS}, as well as our proposed \textit{S-2LS} and \textit{SD-2LS}. For all three methods, we run $k$-means with $K=1024$ clusters on each dataset, which approximates $\sqrt{N}$ for the natural corpora. The same partition and centroids are shared across the three variants, so the only varying factor is the cluster sampling rule (Appendix~\ref{app:figures} reports histograms of cluster~sizes).
SD-2LS uses full-rank covariance matrices $\Sigma_k \in \mathbb{R}^{d \times d}$, so that latency reflects a faithful, non-approximated implementation. 
We further consider \textit{top-$k$ softmax}~\cite{bendada2025exploring}, which restricts the softmax to the $k=1000$ most similar items, implemented via Faiss IVFFlat~\cite{douze2024faiss}, and \textit{hierarchical softmax}~\cite{morin2005hierarchical} as a representative tree-based method. We adopt a recursive binary $k$-means tree to control the memory footprint; details are provided in Appendix~\ref{app:figures}. Across all methods, we vary the temperature parameter $\tau \in \{0.05, 0.1, 0.2\}$ to consider different sharpness levels of the target softmax distribution.

\paragraph{Evaluation Protocol} For each dataset, we sample $n_q=1000$ embeddings and use them as queries. For each query, we compare the item sampling probabilities of exact softmax to those of the five approximate methods for each temperature.  We report three evaluation metrics. First, the \textit{sampling ratio} from Section~\ref{sec:bias} captures biases compared to softmax. Second, the \textit{sampling fidelity} measures how closely each approximate method matches softmax, and is quantified by the Kullback-Leibler (KL) divergence \cite{kullback1951information} $\mathrm{KL}(p_{\text{approx}} \,\|\, p_{\text{exact}})$, averaged over queries. KL is computed in the approximate-to-exact direction, which remains finite when the approximate distribution assigns zero probability to some items, as in top-$k$~softmax. 
Finally, we compute the \textit{per-query latency}, measured in milliseconds and averaged over queries and the three temperatures, as latency is effectively $\tau$-invariant for all methods. All experiments are run single-threaded on CPU with sequential queries, reflecting the inference regime in which 2LS-style samplers are  typically deployed. Our Python code and data are released on GitHub to reproduce all results: \url{https://github.com/bendadaw/neurips-2ls}.

\begin{figure}[t]
\centering
\begin{subfigure}[t]{0.31\linewidth}
    \centering
    \includegraphics[width=\linewidth]{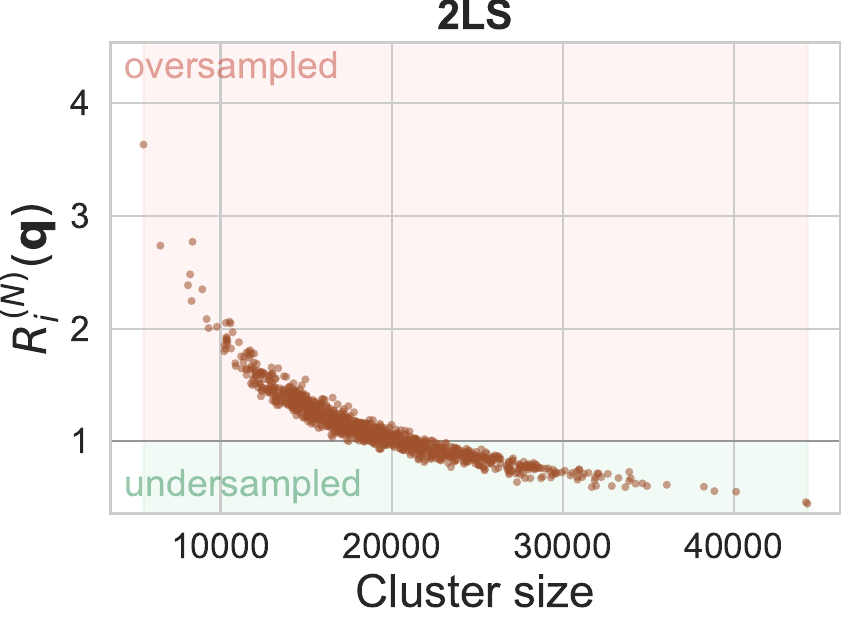}
\end{subfigure}
\hfill
\begin{subfigure}[t]{0.31\linewidth}
    \centering
    \includegraphics[width=\linewidth]{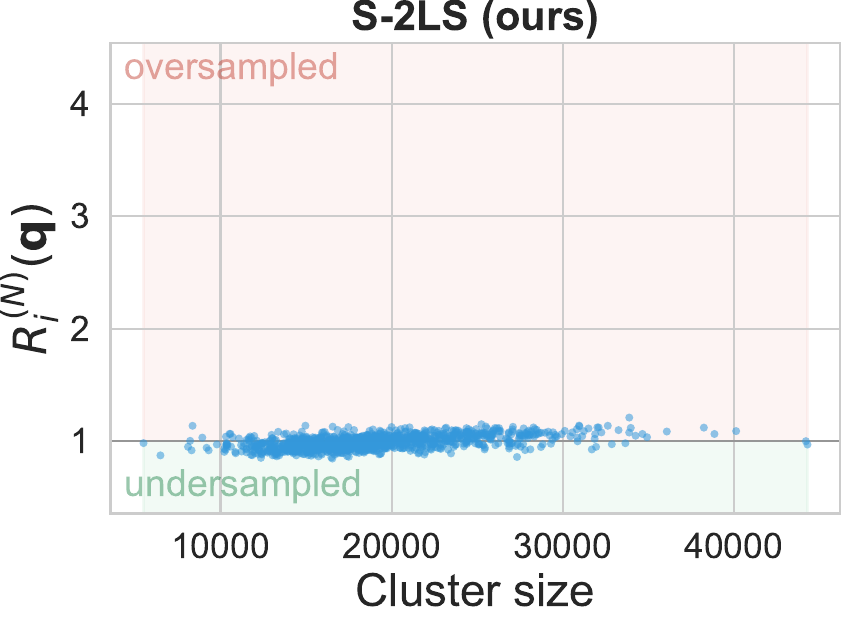}
\end{subfigure}
\hfill
\begin{subfigure}[t]{0.31\linewidth}
    \centering
    \includegraphics[width=\linewidth]{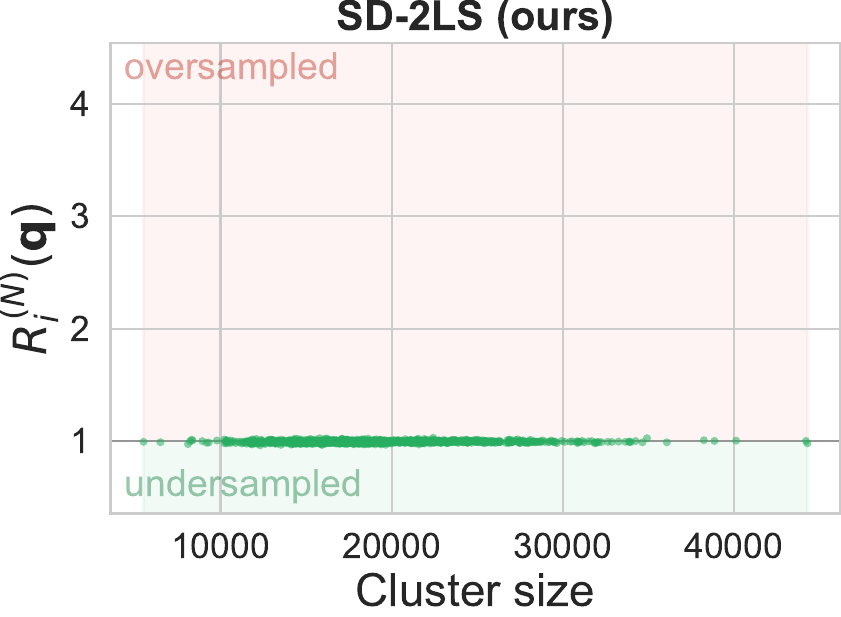}
\end{subfigure}
\caption{Sampling ratio on VK-LSVD ($\tau = 0.1$) as a function of cluster size, averaged over queries. Standard 2LS  undersamples large clusters and oversamples small ones relative to exact softmax ($R_i^{(N)}(q) \neq 1$, p-value $p < 0.01$), while S-2LS and SD-2LS correct these biases.}
\label{fig:section5-illustration1}
\end{figure}

\begin{figure}[t]
\centering
\begin{subfigure}[t]{0.325\linewidth}
    \centering
    \includegraphics[width=\linewidth]{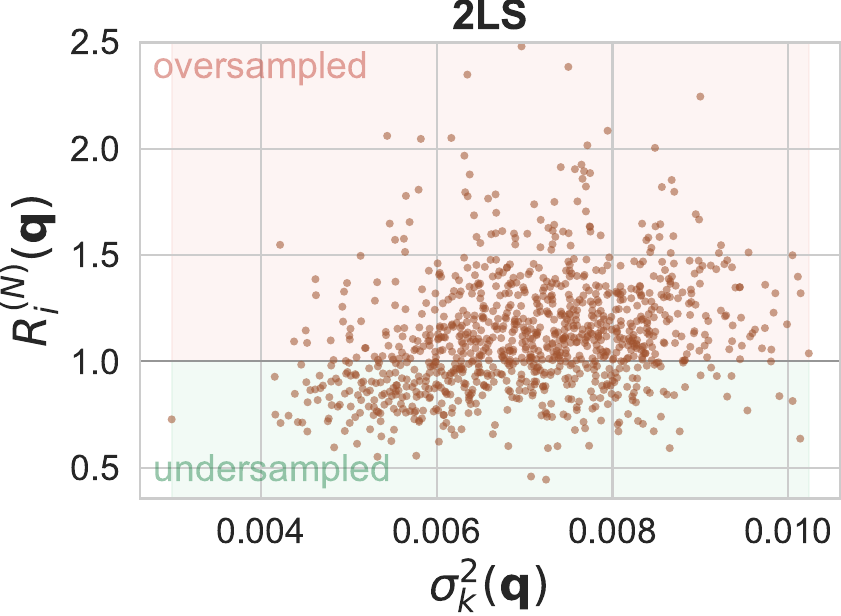}
\end{subfigure}
\hfill
\begin{subfigure}[t]{0.325\linewidth}
    \centering
    \includegraphics[width=\linewidth]{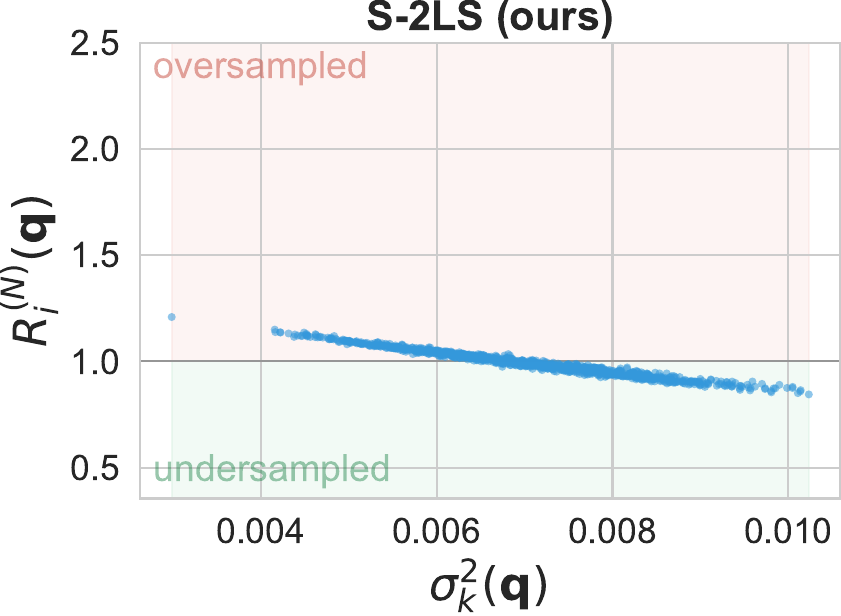}
\end{subfigure}
\hfill
\begin{subfigure}[t]{0.325\linewidth}
    \centering
    \includegraphics[width=\linewidth]{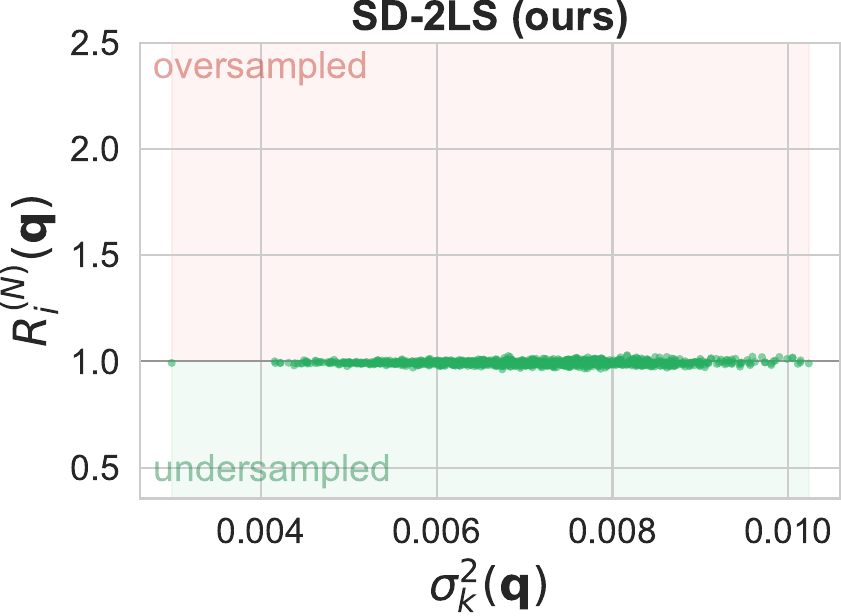}
\end{subfigure}
\caption{Sampling ratio on VK-LSVD ($\tau = 0.1$) as a function of intra-cluster similarity variance, averaged over queries. The effect is entangled with cluster-size bias for 2LS; S-2LS corrects the cluster-size bias but not the variance bias, isolating the latter and revealing that high- (resp., low-) variance clusters are undersampled (resp., oversampled) relative to softmax ($R_i^{(N)}(q) \neq 1$,~$p < 0.01$).}
\label{fig:section5-illustration2}
\end{figure}
\begin{figure}[th!]
\centering
\begin{subfigure}[t]{0.305\linewidth}
    \centering
    \includegraphics[width=\linewidth]{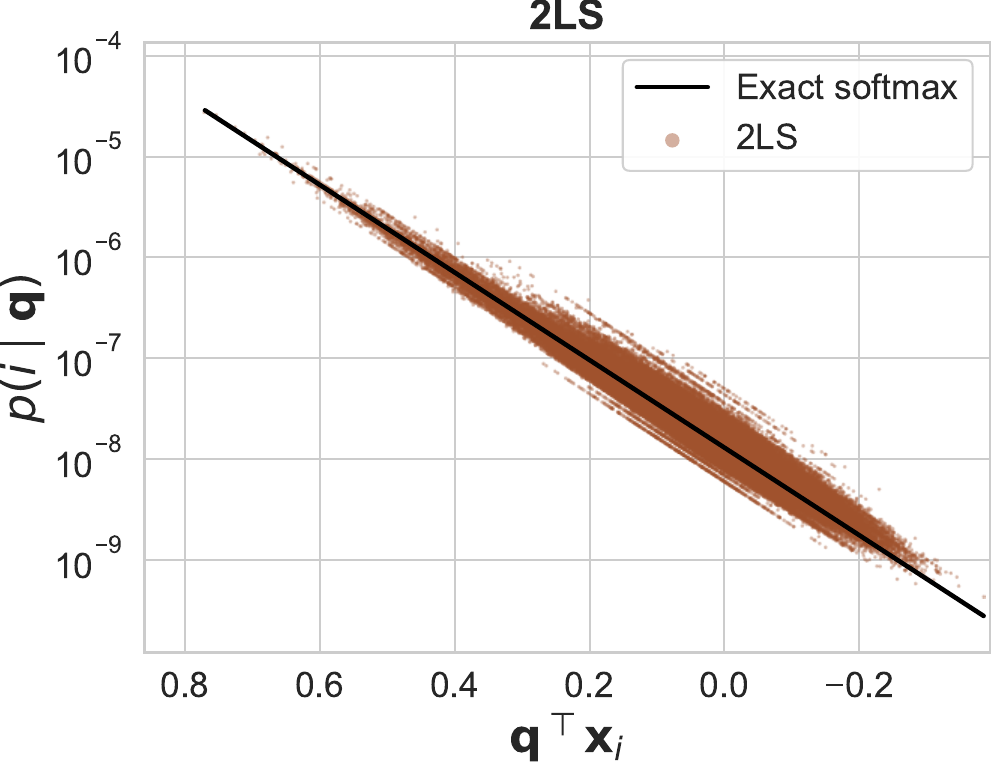}
\end{subfigure}
\hfill
\begin{subfigure}[t]{0.305\linewidth}
    \centering
    \includegraphics[width=\linewidth]{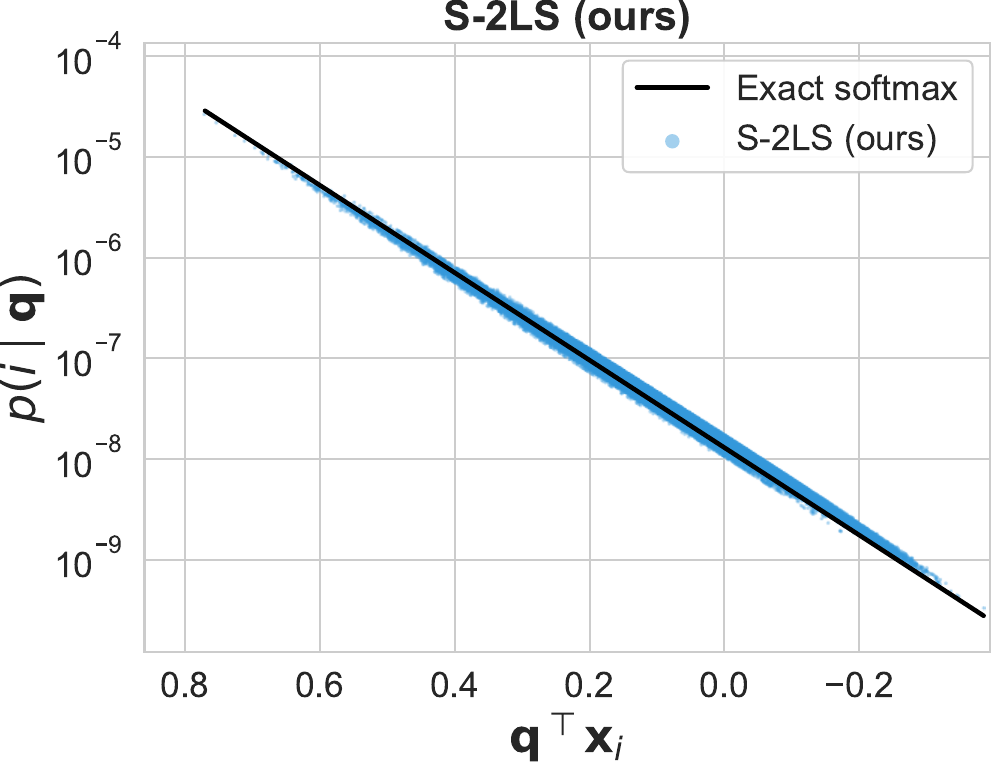}
\end{subfigure}
\hfill
\begin{subfigure}[t]{0.305\linewidth}
    \centering
    \includegraphics[width=\linewidth]{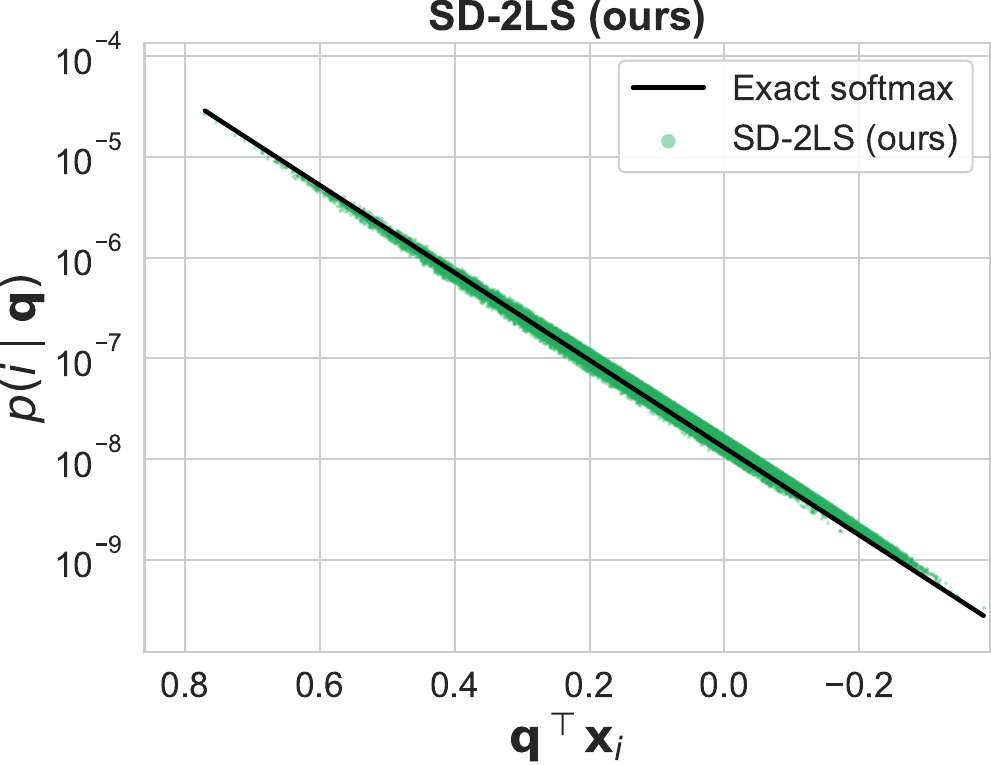}
\end{subfigure}
\caption{Sampling probability on VK-LSVD ($\tau = 0.1$) per item as a function of query-item dot-product similarity for a fixed query. Standard 2LS assigns widely different probabilities to items with identical similarity to the query, while S-2LS and SD-2LS markedly reduce this dispersion.}
\label{fig:section5-illustration3}
\end{figure}

\subsection{Results and Discussion}
\label{subsec:results}

\paragraph{Q1: 2LS biases are substantial in practice}
Our theoretical analysis in Section~\ref{sec:bias}  demonstrates that 2LS misweights clusters by ignoring both cluster-size imbalance and intra-cluster dispersion. Figure~\ref{fig:section5-illustration1} confirms empirically on VK-LSVD that 2LS significantly undersamples large clusters and oversamples small ones relative to exact softmax. 
In addition, Figure~\ref{fig:section5-illustration2} illustrates the impact of intra-cluster similarity variance. While this effect is entangled with the (stronger) cluster-size bias for 2LS, it becomes clearer in the S-2LS figure, which corrects the cluster-size bias but not the intra-cluster variance bias, thereby isolating the latter. Results confirms that 2LS (and S-2LS) undersample clusters with high variability in item-query similarity and oversample those with low variability. Similar results are consistently observed on the four other datasets and reported in Appendix~\ref{app:figures}.
These observations empirically validate the theoretical predictions of Section~\ref{sec:bias} and show that the resulting biases are clearly observable and practically significant across all datasets.

\paragraph{Q2: S-2LS and SD-2LS effectively correct 2LS biases}
Figure~\ref{fig:section5-illustration1} shows on VK-LSVD that S-2LS and SD-2LS effectively correct the cluster-size bias, with sampling ratios tightly concentrated around $R_i^{(N)}(q) = 1$. 
Figure~\ref{fig:section5-illustration2} further shows that, while S-2LS still exhibits a residual bias with respect to within-cluster dispersion, SD-2LS additionally corrects this second source of distortion. Similar patterns are observed across the four remaining datasets and reported in Appendix~\ref{app:figures}.
These results confirm that the proposed corrections substantially reduce the biases of 2LS and yield distributions that more closely match exact softmax. This is further supported by Table~\ref{tab:kl-all-tau}, where S-2LS and SD-2LS consistently achieve higher sampling fidelity than 2LS, as discussed in the next paragraph.

\paragraph{Q3: S-2LS and SD-2LS achieve a superior fidelity--latency trade-off}

Table~\ref{tab:kl-all-tau} reports sampling fidelity across all datasets and temperatures. SD-2LS consistently achieves the lowest KL on all natural corpora (VK-LSVD, YAMBDA, GloVe-100) across all temperatures. The gains over standard 2LS are substantial, ranging from $3$--$5\times$ at $\tau=0.05$ to one to four orders of magnitude at $\tau \in \{0.1, 0.2\}$. S-2LS is a consistent second best, already reducing KL by one to two orders of magnitude in the low-temperature regime most relevant for recommendation. In contrast, top-$k$ truncation exhibits large errors even at $k=1000$, and hierarchical softmax remains significantly less accurate, particularly on YAMBDA, likely due to accumulated routing errors along the tree.

The gap between methods widens with temperature. As $\tau$ increases, the softmax distribution flattens, and accurate estimation of cluster mass becomes the dominant challenge. In this regime, the dispersion correction in SD-2LS becomes critical, explaining the large performance gains observed at $\tau \in \{0.1, 0.2\}$. Synthetic datasets further confirm this mechanism: on Synth-balanced, where cluster sizes are nearly uniform, the advantage of the corrections is limited and top-$k$ can perform well at low temperature; on Synth-unbalanced, which mirrors the heavy-tailed structure of natural corpora, SD-2LS consistently dominates and the gap to 2LS reaches an order of magnitude.

These improvements come at minimal additional cost, as shown in Table~\ref{tab:latency} from Appendix~\ref{app:figures}. S-2LS matches the latency of 2LS within measurement noise, and therefore Pareto-dominates 2LS (same cost, lower KL). SD-2LS incurs a moderate overhead due to the computation of per-cluster quadratic forms, but this cost is efficiently amortized in practice, resulting in only a $1.2$--$1.8\times$ slowdown on higher-dimensional datasets and up to $\sim3\times$ on lower-dimensional ones. Even in this regime, SD-2LS remains within a small constant factor of the fastest approximate methods while delivering KL reductions of several orders of magnitude. Overall, S-2LS and SD-2LS provide a strictly better fidelity-latency trade-off, making them practical drop-in replacements for 2LS in large-scale settings.

\begin{table}[t]
\centering
\caption{Sampling fidelity KL$(p_{\text{approx}} \,\|\, p_{\text{exact}})$ relative to exact softmax, averaged over $n_q=1000$ queries with $95\%$ confidence intervals. Best results are in bold, and second-best are underlined.}
\label{tab:kl-all-tau}
\resizebox{\textwidth}{!}{%
\begin{tabular}{c lccccc}
\toprule
\textbf{$\tau$} & \textbf{Dataset} & \textbf{Top-$k$} & \textbf{2LS} & \textbf{Hierarchical} & \textbf{S-2LS (ours)} & \textbf{SD-2LS (ours)} \\
\midrule

\multirow{5}{*}{0.05} & GloVe-100 & 1.9137 $\pm$ 0.0516 & 0.0961 $\pm$ 0.0110 & 0.5231 $\pm$ 0.0272 & \underline{0.0711 $\pm$ 0.0115} & \textbf{0.0288 $\pm$ 0.0076} \\
 & VK-LSVD & 2.4902 $\pm$ 0.0442 & \underline{0.0541 $\pm$ 0.0023} & 0.7610 $\pm$ 0.0253 & 0.0551 $\pm$ 0.0023 & \textbf{0.0129 $\pm$ 0.0008} \\
 & YAMBDA & 1.7466 $\pm$ 0.0285 & \underline{0.0485 $\pm$ 0.0018} & 1.4484 $\pm$ 0.0563 & 0.0551 $\pm$ 0.0016 & \textbf{0.0177 $\pm$ 0.0056} \\
 & Synth-balanced & \textbf{0.0003 $\pm$ 0.0001} & 0.5968 $\pm$ 0.1051 & 9.4151 $\pm$ 0.0980 & \underline{0.3929 $\pm$ 0.0801} & 0.6586 $\pm$ 0.1403 \\
 & Synth-unbalanced & \textbf{0.0090 $\pm$ 0.0020} & 0.8959 $\pm$ 0.1504 & 10.3033 $\pm$ 0.1216 & 0.7096 $\pm$ 0.1279 & \underline{0.1740 $\pm$ 0.0584} \\

\midrule

\multirow{5}{*}{0.1} & GloVe-100 & 3.9907 $\pm$ 0.0395 & 0.0615 $\pm$ 0.0003 & 0.2897 $\pm$ 0.0018 & \underline{0.0022 $\pm$ 0.0002} & \textbf{0.0001 $\pm$ 0.0000} \\
 & VK-LSVD & 5.2351 $\pm$ 0.0334 & 0.0297 $\pm$ 0.0003 & 0.3425 $\pm$ 0.0027 & \underline{0.0051 $\pm$ 0.0001} & \textbf{0.0002 $\pm$ 0.0000} \\
 & YAMBDA & 3.5266 $\pm$ 0.0292 & 0.0300 $\pm$ 0.0007 & 0.6049 $\pm$ 0.0154 & \underline{0.0098 $\pm$ 0.0003} & \textbf{0.0005 $\pm$ 0.0000} \\
 & Synth-balanced & \underline{0.1580 $\pm$ 0.0087} & 0.3518 $\pm$ 0.0436 & 2.6158 $\pm$ 0.0438 & 0.2334 $\pm$ 0.0327 & \textbf{0.0293 $\pm$ 0.0042} \\
 & Synth-unbalanced & 0.2318 $\pm$ 0.0116 & 0.2858 $\pm$ 0.0391 & 3.1780 $\pm$ 0.0546 & \underline{0.1728 $\pm$ 0.0295} & \textbf{0.0314 $\pm$ 0.0070} \\

\midrule

\multirow{5}{*}{0.2} & GloVe-100 & 5.4103 $\pm$ 0.0217 & 0.0677 $\pm$ 0.0001 & 0.3007 $\pm$ 0.0005 & \underline{0.0001 $\pm$ 0.0000} & \textbf{1.2e-06 $\pm$ 9e-08} \\
 & VK-LSVD & 7.3333 $\pm$ 0.0174 & 0.0337 $\pm$ 0.0001 & 0.3153 $\pm$ 0.0007 & \underline{0.0003 $\pm$ 0.0000} & \textbf{2.7e-06 $\pm$ 1e-07} \\
 & YAMBDA & 5.6988 $\pm$ 0.0202 & 0.0388 $\pm$ 0.0003 & 0.3263 $\pm$ 0.0020 & \underline{0.0008 $\pm$ 0.0000} & \textbf{5.3e-06 $\pm$ 2e-07} \\
 & Synth-balanced & 2.2841 $\pm$ 0.0261 & 0.1296 $\pm$ 0.0026 & 0.6342 $\pm$ 0.0029 & \underline{0.0117 $\pm$ 0.0015} & \textbf{0.0083 $\pm$ 0.0021} \\
 & Synth-unbalanced & 2.2393 $\pm$ 0.0345 & 0.1935 $\pm$ 0.0057 & 0.9311 $\pm$ 0.0070 & \underline{0.0059 $\pm$ 0.0009} & \textbf{0.0023 $\pm$ 0.0011} \\

\bottomrule
\end{tabular}%
}
\end{table}

\section{Conclusion}
\label{sec:conclusion}

In this paper, we showed that 2LS sampling introduces systematic biases due to misweighting clusters. We characterized these biases theoretically and demonstrated that they are substantial in practice across multiple large-scale datasets. We proposed two corrections, S-2LS and SD-2LS, which preserve the computational efficiency of 2LS while correcting its biases and improving softmax sampling fidelity. Extensive theoretical and experimental results show that S-2LS consistently improves upon 2LS at no cost, while SD-2LS further reduces approximation error with only moderate overhead.
These results support clear practical recommendations: S-2LS should replace standard 2LS unconditionally, as it incurs no additional cost and consistently improves sampling fidelity. SD-2LS should be preferred when intra-cluster variance is non-negligible and the additional computational cost is acceptable, as in many large-scale recommendation settings where $d \ll N$.
Finally, our analysis suggests directions for future research. In particular, as SD-2LS relies on a second-order approximation of the similarity distribution, extending our method to account for higher-order moments could further improve sampling accuracy for embedding distributions with high skewness~or~kurtosis.


\clearpage

\bibliographystyle{plainnat} 
\bibliography{references}


\appendix
\clearpage
\section*{Appendix}

This appendix supplements the article "\textit{Two-Level Softmax Sampling Done Right:
Correcting Bias from Size Imbalance and Dispersion}" and is organized as follows:
\begin{itemize}
    \item Appendix \ref{appendix:proofs-section3} presents detailed proofs for all theoretical results presented in Section \ref{sec:bias}.
        \item Appendix \ref{appendix:proofs-section4} reports detailed proofs for all theoretical results presented in Section \ref{sec:sd2ls}.
        \item Appendix \ref{app:figures} provides additional results and technical details complementing the experimental analysis in Section~\ref{sec:experiments}.
\end{itemize}

\section{Proofs for Section~\ref{sec:bias}: Bias Characterization of 2LS Sampling}
\label{appendix:proofs-section3}

\subsection{Proof of Proposition~\ref{prop:cluster}}

Equation~\eqref{eq:ratio} in Section~\ref{sec:bias} introduced the sampling ratio 
$R_i^{(N)}(q) = p_{\text{2LS}}(i \mid q) / p(i \mid q)$. Substituting the expressions of the sampling probabilities $p_{\text{2LS}}(i \mid q)$ and  $p(i \mid q)$ from Section~\ref{sec:preliminaries}, we obtain
\begin{align}
R_i^{(N)}(q) 
&= \frac{\frac{\exp(q^\top \mu_k / \tau)}
{\sum_{k'=1}^K \exp(q^\top \mu_{k'} / \tau)} 
\cdot 
\frac{\exp(q^\top x_i / \tau)}
{\sum_{j \in \mathcal{C}_k} \exp(q^\top x_j / \tau)}}
{\frac{\exp(q^\top x_i / \tau)}
{\sum_{j=1}^N \exp(q^\top x_j / \tau)}} \\
&= \frac{\exp(q^\top \mu_k / \tau)}{\sum_{k'=1}^K \exp(q^\top \mu_{k'} / \tau)}
\cdot
\frac{\sum_{j=1}^N \exp(q^\top x_j / \tau)}{\sum_{j \in \mathcal{C}_k} \exp(q^\top x_j / \tau)}.
\end{align}

As the $\exp(q^\top x_k / \tau)$ terms cancel out, the resulting expression depends only on the cluster $\mathcal{C}_k$ containing $i$, and not on the dot product similarity $q^\top x_i$ directly. Consequently, $R_i^{(N)}(q)$ is constant across all items within the same cluster $\mathcal{C}_k$, which proves the proposition. \QED

\subsection{Proof of Proposition \ref{prop:main}}

From the previous proof, we have:
\begin{equation}
R_i^{(N)}(q)
=
\frac{\exp(q^\top \mu_{k} / \tau)}
{\sum_{k'=1}^K \exp(q^\top \mu_{k'} / \tau)}
\cdot
\frac{\sum_{j=1}^N \exp(q^\top x_j/ \tau)}
{\sum_{j \in \mathcal C_{k}} \exp(q^\top x_j/ \tau)}.
\label{eq:ratio-proof-start}
\end{equation}

We rewrite the second factor in a form amenable to asymptotic analysis. To begin with,
\begin{equation}
\sum_{j \in \mathcal C_{k}} \exp(q^\top x_j/ \tau)
=
|\mathcal C_{k}|
\cdot
\frac{1}{|\mathcal C_{k}|}
\sum_{j \in \mathcal C_{k}} \exp(q^\top x_j/ \tau).
\end{equation}
Also,
\begin{equation}
\sum_{j=1}^N \exp(q^\top x_j/ \tau)
=
N \cdot \frac{1}{N}\sum_{j=1}^N \exp(q^\top x_j/ \tau).
\end{equation}
Substituting these identities into Equation \eqref{eq:ratio-proof-start}, we obtain
\begin{equation}
R_i^{(N)}(q)
=
\frac{\exp(q^\top \mu_{k}/ \tau)}
{\sum_{k'=1}^K \exp(q^\top \mu_{k'}/ \tau)}
\cdot
\frac{
\frac{1}{N}\sum_{j=1}^N \exp(q^\top x_j/ \tau)
}{
\frac{|\mathcal C_{k}|}{N}
\cdot
\frac{1}{|\mathcal C_{k}|}
\sum_{j \in \mathcal C_{k}} \exp(q^\top x_j/ \tau)
}.
\label{eq:ratio-proof-rewritten}
\end{equation}

We analyze each factor separately. First, since $x_1, \dots, x_N$ are i.i.d.\ samples from the mixture distribution, the strong law of large numbers yields
\begin{equation}
\frac{1}{N}\sum_{j=1}^N \exp(q^\top x_j/ \tau)
\xrightarrow[N \to \infty]{a.s.}
\mathbb E[\exp(q^\top x/ \tau)].
\label{eq:global-lln}
\end{equation}

Second, for each $k \in \{1,\dots,K\}$, the empirical cluster proportion converges almost surely to the corresponding mixture weight, i.e.,
\begin{equation}
\frac{|\mathcal C_{k}|}{N}
\xrightarrow[N \to \infty]{a.s.}
\pi_{k}.
\label{eq:cluster-proportion}
\end{equation}
 Third,the embeddings from component $\mathcal C_{k}$ are i.i.d. samples from $\mathcal D_{k}$. Hence, by the strong law of large numbers,
\begin{equation}
\frac{1}{|\mathcal C_{k}|}
\sum_{j \in \mathcal C_{k}} \exp(q^\top x_j/ \tau)
\xrightarrow[N \to \infty]{a.s.}
\mathbb E_{x\sim \mathcal D_{k}}[\exp(q^\top x/ \tau)].
\label{eq:within-cluster-lln}
\end{equation}

Combining Equation~\eqref{eq:ratio-proof-rewritten}, Equation~\eqref{eq:global-lln}, Equation~\eqref{eq:cluster-proportion}, and Equation~\eqref{eq:within-cluster-lln}, we obtain
\begin{equation}
R_i^{(N)}(q)
\xrightarrow[N \to \infty]{a.s.}
\frac{\exp(q^\top \mu_{k}/\tau)}
{\sum_{k'=1}^K \exp(q^\top \mu_{k'}/\tau)}
\cdot
\frac{\mathbb E[\exp(q^\top x/\tau)]}
{\pi_{k}\, \mathbb E_{x\sim \mathcal D_{k}}[\exp(q^\top x/\tau)]}.
\end{equation}
Defining
\begin{equation}
c(q)
=
\frac{\mathbb E[\exp(q^\top x/\tau)]}
{\sum_{k'=1}^K \exp(q^\top \mu_{k'}/\tau)},
\end{equation}
this limit can be written as
\begin{equation}
R_i^{(N)}(q)
\xrightarrow[N \to \infty]{a.s.}
c(q)\,
\frac{\exp(q^\top \mu_{k}/\tau)}
{\pi_{k}\, \mathbb E_{x\sim \mathcal D_{k}}[\exp(q^\top x/\tau)]}
=
R_{k}^{(\infty)}(q).
\end{equation}
This concludes the proof. \QED

\subsection{Proof of Proposition~\ref{corollary:gaussian}}

By Proposition~\ref{prop:main}, for any $k \in \{1,\dots,K\}$,
\begin{equation}
R_k^{(\infty)}(q)
=
c(q)\,
\frac{\exp(q^\top \mu_k/\tau)}
{\pi_k\,\mathbb E_{x\sim \mathcal D_k}[\exp(q^\top x/\tau)]}.
\label{eq:cor-start}
\end{equation}

Under a Gaussian approximation, the scalar projection $q^\top x/\tau$ with $x \sim \mathcal D_k$ is modeled as a Gaussian random variable with mean $q^\top \mu_k/\tau$ and variance $\sigma_k^2(q)$. Using the moment generating function of a Gaussian distribution \cite{casella2024statistical}, we have
\begin{equation}
\mathbb E_{x\sim \mathcal D_k}[\exp(q^\top x/\tau)]
=
\exp\!\left(q^\top \mu_k/\tau+\tfrac12 \sigma_k^2(q)\right).
\label{eq:gauss-mgf}
\end{equation}

Substituting Equation \eqref{eq:gauss-mgf} into Equation \eqref{eq:cor-start}, we obtain
\begin{equation}
R_k^{(\infty)}(q)
=
c(q)\,
\frac{\exp(q^\top \mu_k/\tau)}
{\pi_k\,\exp\!\left(q^\top \mu_k/\tau+\tfrac12 \sigma_k^2(q)\right)}.
\end{equation}
Canceling the factor $\exp(q^\top \mu_k/\tau)$ gives
\begin{equation}
R_k^{(\infty)}(q)
=
c(q)\,
\frac{1}{\pi_k\,\exp\!\left(\tfrac12 \sigma_k^2(q)\right)}.
\end{equation}
Since $c(q)$ does not depend on $k$, this implies
\begin{equation}
R_k^{(\infty)}(q)
\propto
\frac{1}{\pi_k\,\exp\!\left(\tfrac12 \sigma_k^2(q)\right)}.
\end{equation} \QED

\section{Proofs for Section \ref{sec:sd2ls}: Size- and Dispersion-Corrected 2LS Sampling}
\label{appendix:proofs-section4}

\subsection{Proof of Proposition \ref{prop:first_order_ratio}}

The proof follows the steps of that of Proposition~\ref{prop:main}, with the difference that the cluster-level distribution is now weighted by $|\mathcal C_k|$. Accordingly, for any $i \in \mathcal C_k$,
\begin{equation}
R_{i,\text{S-2LS}}^{(N)}(q)
=
\frac{|\mathcal C_k| \exp(q^\top \mu_k/\tau)}
{\sum_{k'=1}^K |\mathcal C_{k'}| \exp(q^\top \mu_{k'}/\tau)}
\cdot
\frac{\sum_{j=1}^N \exp(q^\top x_j/\tau)}
{\sum_{j \in \mathcal C_k} \exp(q^\top x_j/\tau)}.
\end{equation}
Dividing the numerator and denominator of the first factor by $N$, and rewriting the denominator of the second factor as in Proposition~\ref{prop:main}, we obtain
\begin{equation}
R_{i,\text{S-2LS}}^{(N)}(q)
=
\frac{\frac{|\mathcal C_k|}{N} \exp(q^\top \mu_k/\tau)}
{\sum_{k'=1}^K \frac{|\mathcal C_{k'}|}{N} \exp(q^\top \mu_{k'}/\tau)}
\cdot
\frac{\frac{1}{N}\sum_{j=1}^N \exp(q^\top x_j/\tau)}
{\frac{|\mathcal C_k|}{N}
\cdot
\frac{1}{|\mathcal C_k|}
\sum_{j \in \mathcal C_k} \exp(q^\top x_j/\tau)}.
\end{equation}
Applying the same almost-sure limits as in Proposition~\ref{prop:main}, namely,
\begin{equation}
\frac{|\mathcal C_k|}{N} \xrightarrow[N \to \infty]{a.s.} \pi_k,
\end{equation}
\begin{equation}
\frac{1}{N}\sum_{j=1}^N \exp(q^\top x_j/\tau)
\xrightarrow[N \to \infty]{a.s.}
\mathbb E_{x\sim \sum_{k'=1}^K \pi_{k'}\mathcal D_{k'}}[\exp(q^\top x/\tau)],
\end{equation}
and
\begin{equation}
\frac{1}{|\mathcal C_k|}
\sum_{j \in \mathcal C_k} \exp(q^\top x_j/\tau)
\xrightarrow[N \to \infty]{a.s.}
\mathbb E_{x\sim \mathcal D_k}[\exp(q^\top x/\tau)],
\end{equation}
yields
\begin{equation}
R_{i,\text{S-2LS}}^{(N)}(q)
\xrightarrow[N \to \infty]{a.s.}
c_{\text{S-2LS}}(q)\,
\frac{\exp(q^\top \mu_k/\tau)}
{\mathbb E_{x\sim \mathcal D_k}[\exp(q^\top x)/\tau]},
\end{equation}
with
\begin{equation}
c_{\text{S-2LS}}(q)
=
\frac{
\mathbb E_{x\sim \sum_{k'=1}^K \pi_{k'}\mathcal D_{k'}}[\exp(q^\top x/\tau)]
}{
\sum_{k'=1}^K \pi_{k'} \exp(q^\top \mu_{k'}/\tau)
}. 
\end{equation} \QED

\subsection{Proof of Proposition \ref{prop:second_order_ratio}}

Compared to previous proofs, the cluster-level score now includes both the cluster cardinality and the dispersion correction. For any $i \in \mathcal C_k$, we have
\begin{equation}
R_{i,\text{SD-2LS}}^{(N)}(q)
=
\frac{|\mathcal C_k| \exp\!\left(q^\top \mu_k/\tau + \tfrac12 \sigma_k^2(q)\right)}
{\sum_{k'=1}^K |\mathcal C_{k'}| \exp\!\left(q^\top \mu_{k'}/\tau + \tfrac12 \sigma_{k'}^2(q)\right)}
\cdot
\frac{\sum_{j=1}^N \exp(q^\top x_j/\tau)}
{\sum_{j \in \mathcal C_k} \exp(q^\top x_j/\tau)}.
\end{equation}
Applying the same asymptotic argument as in Proposition~\ref{prop:main} and \ref{prop:first_order_ratio}, we obtain
\begin{equation}
R_{i,\text{SD-2LS}}^{(N)}(q)
\xrightarrow[N \to \infty]{a.s.}
c_{\text{SD-2LS}}(q)\,
\frac{\exp\!\left(q^\top \mu_k/\tau + \tfrac12 \sigma_k^2(q)\right)}
{\mathbb E_{x\sim \mathcal D_k}[\exp(q^\top x/\tau)]},
\end{equation}
where
\begin{equation}
c_{\text{SD-2LS}}(q)
=
\frac{
\mathbb E_{x\sim \sum_{k'=1}^K \pi_{k'}\mathcal D_{k'}}[\exp(q^\top x/\tau)]
}{
\sum_{k'=1}^K \pi_{k'} \exp\!\left(q^\top \mu_{k'}/\tau + \tfrac12 \sigma_{k'}^2(q)\right)
},
\end{equation}
which is the result stated in Proposition \ref{prop:second_order_ratio}. \QED

\subsection{Proof of Proposition \ref{prop:second_order_ratio_gaussian}}

In the proof of Proposition \ref{prop:second_order_ratio}, we have obtained
\begin{equation}
R_{i,\text{SD-2LS}}^{(N)}(q)
\xrightarrow[N \to \infty]{a.s.}
c_{\text{SD-2LS}}(q)\,
\frac{\exp\!\left(q^\top \mu_k/\tau + \tfrac12 \sigma_k^2(q)\right)}
{\mathbb E_{x\sim \mathcal D_k}[\exp(q^\top x/\tau)]},
\end{equation}
where
\begin{equation}
c_{\text{SD-2LS}}(q)
=
\frac{
\mathbb E_{x\sim \sum_{k'=1}^K \pi_{k'}\mathcal D_{k'}}[\exp(q^\top x/\tau)]
}{
\sum_{k'=1}^K \pi_{k'} \exp\!\left(q^\top \mu_{k'}/\tau + \tfrac12 \sigma_{k'}^2(q)\right)
}.
\end{equation}
Under the Gaussian approximation of $q^\top x$ under each component $\mathcal D_k$,
\begin{equation}
\mathbb E_{x\sim \mathcal D_k}[\exp(q^\top x/\tau)]
=
\exp\!\left(q^\top \mu_k/\tau + \tfrac12 \sigma_k^2(q)\right),
\end{equation}
so the cluster-dependent factor cancels and
\begin{equation}
R_{i,\text{SD-2LS}}^{(N)}(q) \xrightarrow[N \to \infty]{a.s.} c_{\text{SD-2LS}}(q).
\end{equation}
Applying the same approximation to the mixture expectation in the definition of $c_{\text{SD-2LS}}(q)$ gives
\begin{equation}
c_{\text{SD-2LS}}(q) = 1,
\end{equation}
hence
\begin{equation}
R_{i,\text{SD-2LS}}^{(N)}(q) \xrightarrow[N \to \infty]{a.s.} 1.
\end{equation}
This concludes the proof. \QED

\section{Additional Experimental Results and Technical Details}
\label{app:figures} 

\subsection{Technical Details on Hierarchical Softmax Implementation}

While the primary goal of our experiments is to compare 2LS with our proposed S-2LS and SD-2LS sampling methods, we also include \emph{hierarchical softmax}~\cite{morin2005hierarchical} in the comparisons of Section~\ref{sec:experiments} as a representative tree-based factorization baseline. The textbook variant uses a fully binary tree with one item per leaf, which requires storing roughly $N$ additional internal-node embeddings on top of the dataset itself, effectively doubling the index memory footprint. At the scale of the recommender corpora considered in experiments ($N$ in the tens of millions), this is impractical.

We therefore adopt an intermediate variant: a recursive binary $k$-means tree in which each internal node stores the mean of its descendant subtree as a routing representative, and recursion terminates at $\ell = 1024$ items per leaf. Within-leaf sampling is performed via an exact softmax over the leaf members. This yields $\mathcal{O}(\log_2(N/\ell))$ routing decisions per query, providing a logarithmic speedup over the two-level case, while keeping memory usage comparable to the 2LS family. We verified that the results are robust to the choice of leaf size over the range $[256, 4096]$.

\subsection{Per-Query Latency Across Sampling Methods}

\begin{table}[h!]
\centering
\caption{Per-query latency in milliseconds, averaged over temperatures $\tau \in \{0.05, 0.1, 0.2\}$, with $K=1024$ clusters and $n_q=1000$ queries, reported as mean $\pm$ 95\% confidence intervals. Speed-up gains are computed relative to the exact softmax sampling method, denoted \textit{Exact}. Best results per dataset are in bold, and second-best are underlined.}
\label{tab:latency}
\resizebox{\textwidth}{!}{%
\begin{tabular}{lcccccc}
\toprule
\textbf{Dataset} & \textbf{Exact} & \textbf{Top-$k$} & \textbf{2LS} & \textbf{Hierarchical} & \textbf{S-2LS (ours)} & \textbf{SD-2LS (ours)} \\
\midrule
GloVe-100 & 10.30 $\pm$ 0.04 & 0.439 $\pm$ 0.004 (23.4$\times$) & \underline{0.235 $\pm$ 0.003 (43.9$\times$)} & \textbf{0.221 $\pm$ 0.001 (46.7$\times$)} & 0.245 $\pm$ 0.003 (42.1$\times$) & 0.722 $\pm$ 0.009 (14.3$\times$) \\
VK-LSVD & 173.0 $\pm$ 5.0 & 2.720 $\pm$ 0.019 (63.6$\times$) & \underline{2.564 $\pm$ 0.728 (67.5$\times$)} & \textbf{1.383 $\pm$ 0.019 (125$\times$)} & 2.662 $\pm$ 0.751 (65.0$\times$) & 2.906 $\pm$ 0.529 (59.5$\times$) \\
YAMBDA & 79.28 $\pm$ 6.39 & 2.013 $\pm$ 0.018 (39.4$\times$) & \underline{1.265 $\pm$ 0.063 (62.7$\times$)} & \textbf{1.185 $\pm$ 0.083 (66.9$\times$)} & 1.660 $\pm$ 0.434 (47.7$\times$) & 3.171 $\pm$ 1.227 (25.0$\times$) \\
Synth-Bal & 8.543 $\pm$ 0.036 & 0.289 $\pm$ 0.002 (29.6$\times$) & 0.222 $\pm$ 0.004 (38.4$\times$) & \underline{0.218 $\pm$ 0.001 (39.3$\times$)} & \textbf{0.213 $\pm$ 0.004 (40.2$\times$)} & 0.690 $\pm$ 0.005 (12.4$\times$) \\
Synth-Unbal & 8.619 $\pm$ 0.032 & 0.317 $\pm$ 0.003 (27.2$\times$) & \underline{0.213 $\pm$ 0.004 (40.5$\times$)} & \textbf{0.211 $\pm$ 0.001 (40.8$\times$)} & 0.214 $\pm$ 0.004 (40.2$\times$) & 0.691 $\pm$ 0.005 (12.5$\times$) \\
\bottomrule
\end{tabular}%
}
\end{table}

\clearpage
\subsection{Cluster Size Distributions Across Datasets}

\begin{figure}[h!]
\centering
\begin{subfigure}{0.325\linewidth}
  \includegraphics[width=\linewidth]{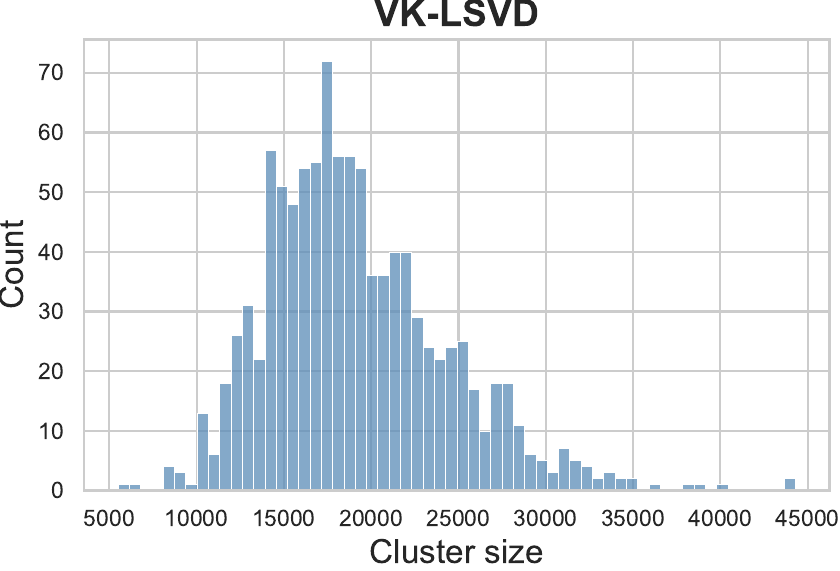}
\end{subfigure}\hfill
\begin{subfigure}{0.325\linewidth}
  \includegraphics[width=\linewidth]{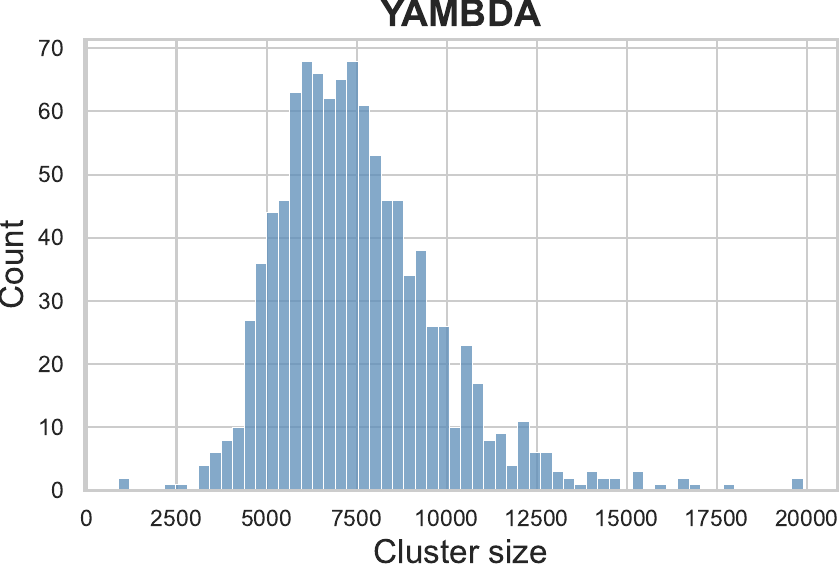}
\end{subfigure}\hfill
\begin{subfigure}{0.325\linewidth}
  \includegraphics[width=\linewidth]{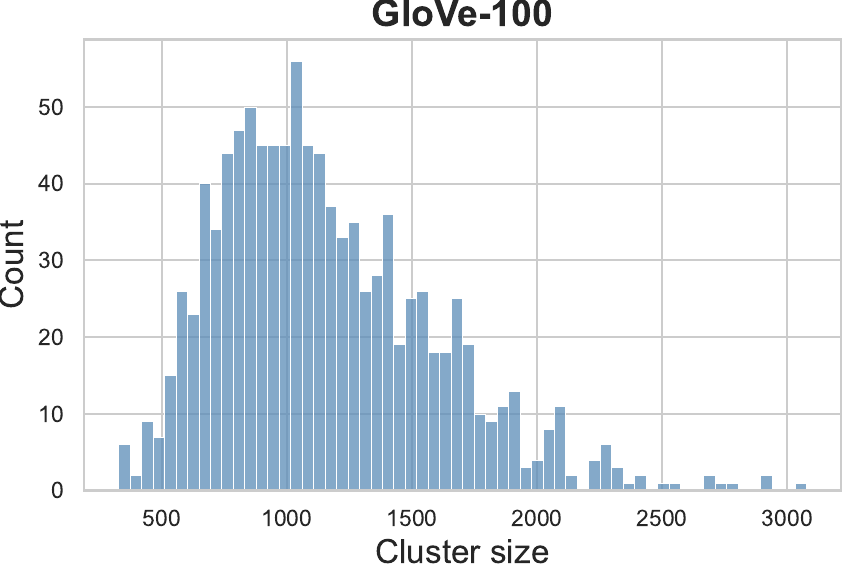}
\end{subfigure}\hfill
\begin{subfigure}{0.325\linewidth}
  \includegraphics[width=\linewidth]{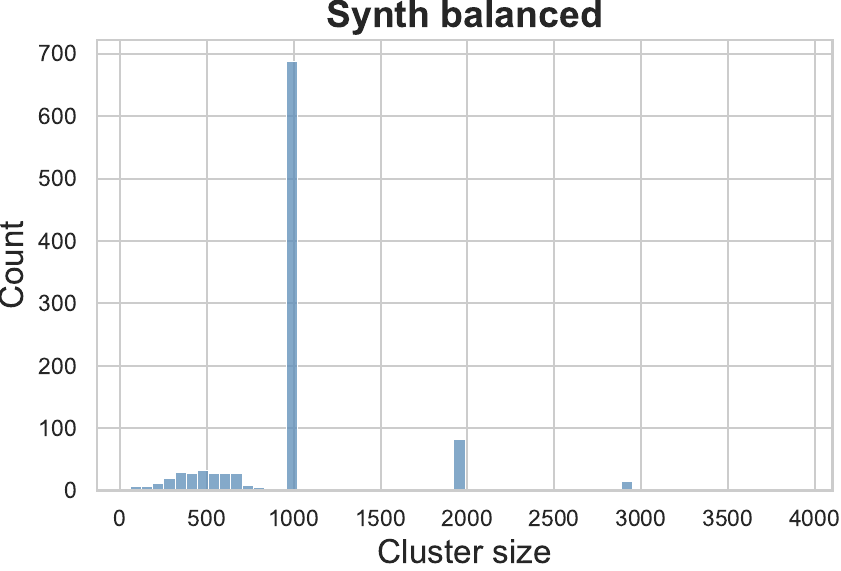}
\end{subfigure} 
\begin{subfigure}{0.325\linewidth}
  \includegraphics[width=\linewidth]{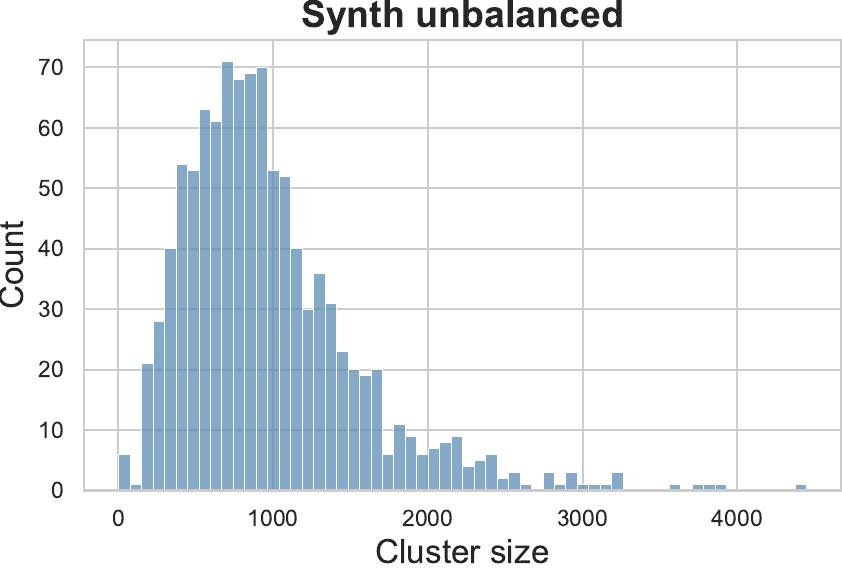}
\end{subfigure}
\caption{Cluster size distributions at $K=1024$ across all five datasets, obtained via $k$-means clustering. The natural recommender corpora (VK-LSVD, YAMBDA) and GloVe-100 exhibit substantial size imbalance, with cluster sizes spanning more than an order of magnitude. The synthetic corpora serve as controlled regimes: Synth-balanced has tightly concentrated cluster sizes, while Synth-unbalanced replicates the heavy-tailed distributions observed in natural corpora. This imbalance is precisely what the size correction in S-2LS and SD-2LS is designed to address. Recall that Synth-balanced was generated from a Gaussian mixture model with an equal number of points per component. However, the $k$-means clustering used in our experiments does not exactly recover the ground-truth components. As a result, some observed clusters have different sizes even on Synth-balanced.}
\label{fig:cluster_size_hist_appendix}
\end{figure}

\subsection{Complete Bias Analysis Across All Datasets}

Figures~\ref{fig:section5-illustration1}, \ref{fig:section5-illustration2}, and \ref{fig:section5-illustration3} in the main paper present experimental results showing that the 2LS biases theoretically characterized in Section~\ref{sec:bias} are observable and substantial in practice, and that the proposed S-2LS and SD-2LS methods effectively correct them.

Specifically, Figure~\ref{fig:section5-illustration1} reports the sampling ratio $R_i^{(N)}(q)$ as a function of cluster size, Figure~\ref{fig:section5-illustration2} reports $R_i^{(N)}(q)$ as a function of intra-cluster similarity variance, and Figure~\ref{fig:section5-illustration3} reports the sampling probability per item as a function of query-item similarity.

Due to space constraints, these figures in the main paper focus on the VK-LSVD dataset. For completeness, we provide in this appendix \textit{the full set of analyses}, including results for the four remaining datasets: YAMBDA, GloVe-100, Synth-Balanced, and Synth-Unbalanced. The results lead to conclusions consistent with those reported in the main paper.

\clearpage


\begin{figure}[t]
\vspace{2.5cm}
\centering
\begin{subfigure}[t]{0.325\linewidth}
    \centering
    \includegraphics[width=\linewidth]{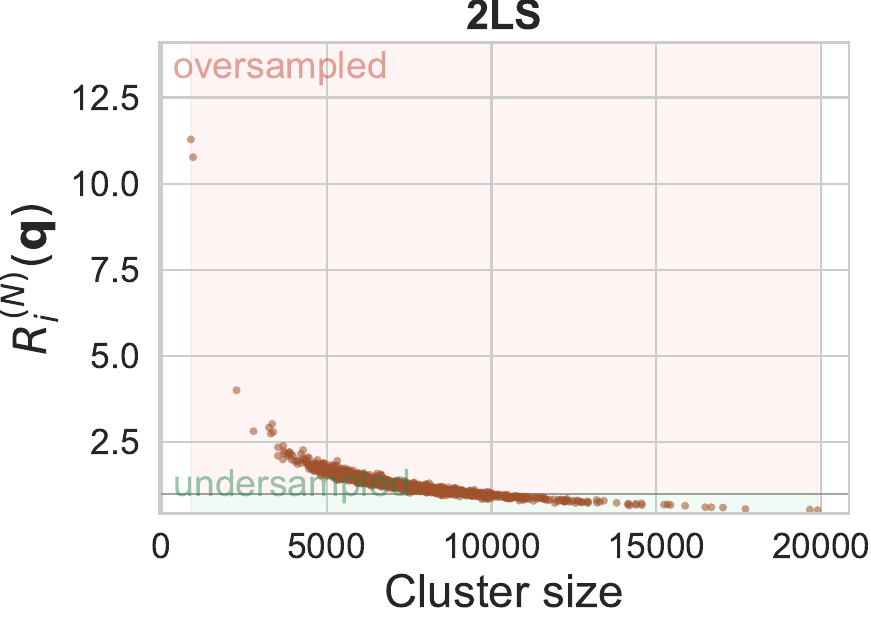}
\end{subfigure}
\hfill
\begin{subfigure}[t]{0.325\linewidth}
    \centering
    \includegraphics[width=\linewidth]{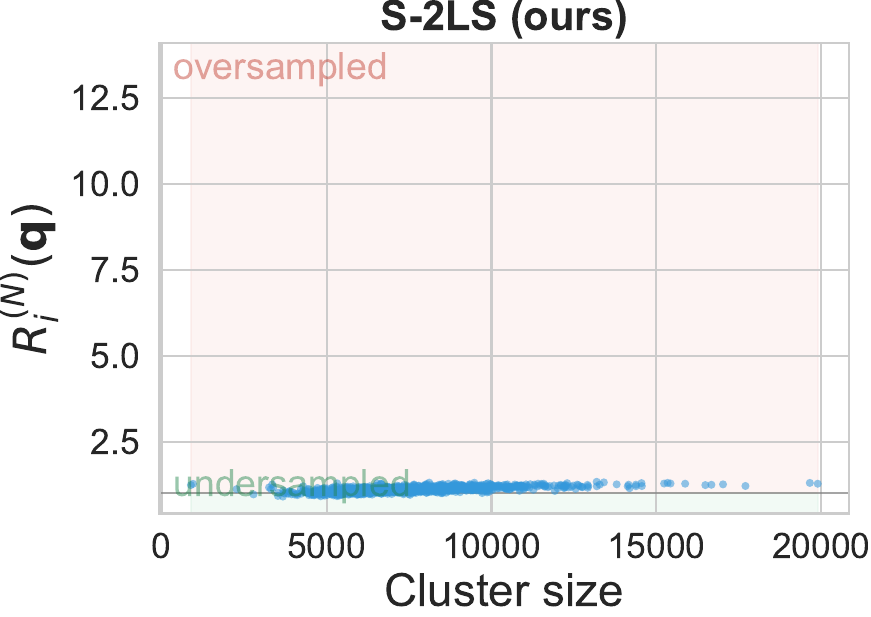}
\end{subfigure}
\hfill
\begin{subfigure}[t]{0.325\linewidth}
    \centering
    \includegraphics[width=\linewidth]{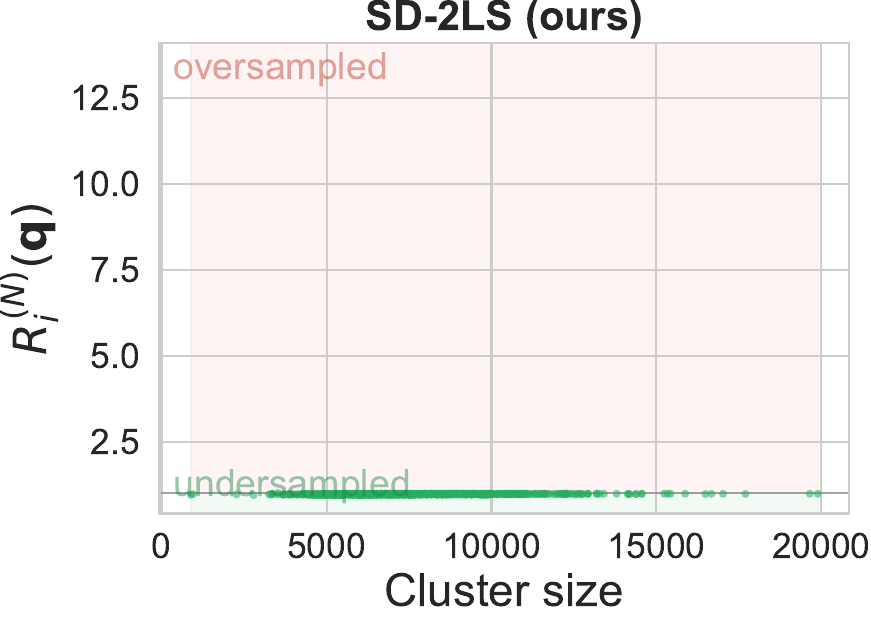}
\end{subfigure}
\caption{Sampling ratio on \textbf{YAMBDA} ($\tau = 0.1$) as a function of cluster size, averaged over queries. Standard 2LS  undersamples large clusters and oversamples small ones relative to exact softmax ($R_i^{(N)}(q) \neq 1$, p-value $p < 0.01$), while S-2LS and SD-2LS correct these biases.}
\end{figure}

\begin{figure}[t]
\centering
\begin{subfigure}[t]{0.32\linewidth}
    \centering
    \includegraphics[width=\linewidth]{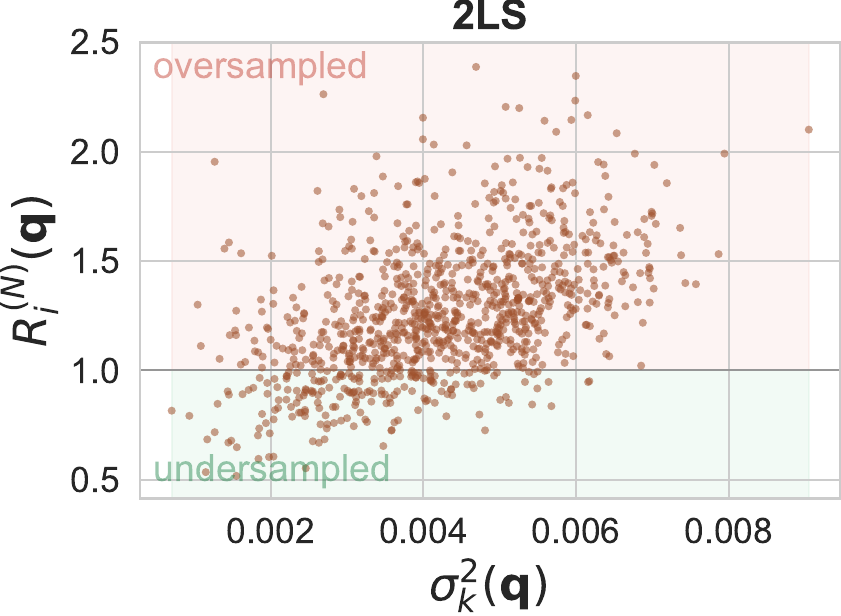}
\end{subfigure}
\hfill
\begin{subfigure}[t]{0.32\linewidth}
    \centering
    \includegraphics[width=\linewidth]{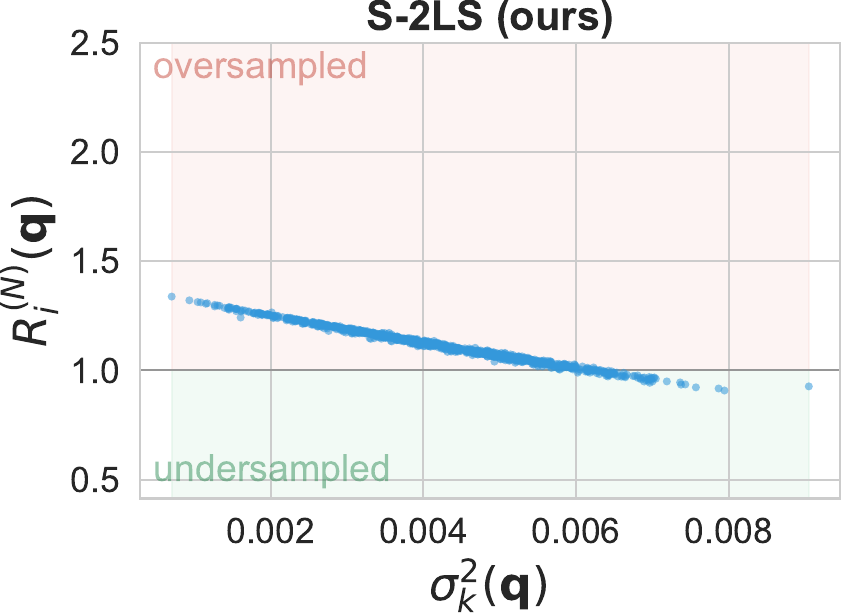}
\end{subfigure}
\hfill
\begin{subfigure}[t]{0.32\linewidth}
    \centering
    \includegraphics[width=\linewidth]{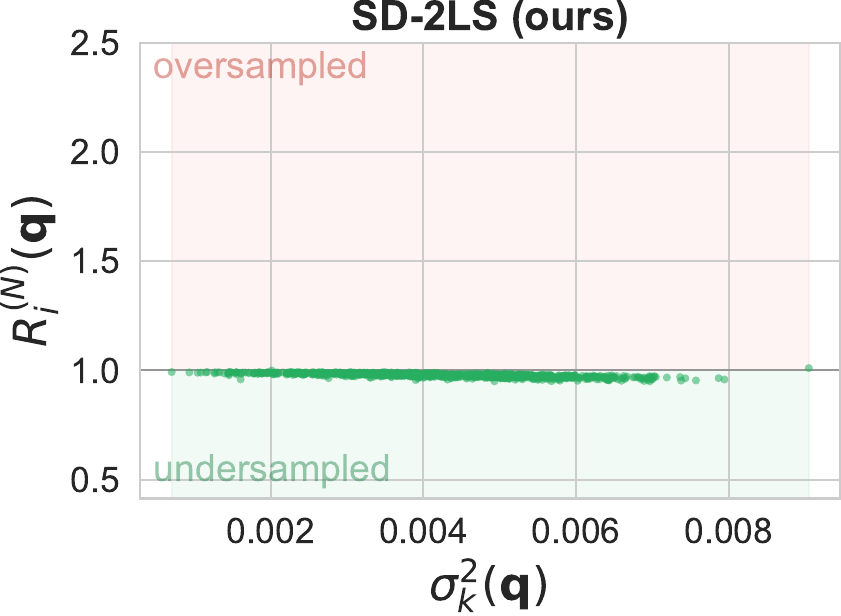}
\end{subfigure}
\caption{Sampling ratio on \textbf{YAMBDA} ($\tau = 0.1$) as a function of intra-cluster similarity variance, averaged over queries. The effect is entangled with cluster-size bias for 2LS; S-2LS corrects the cluster-size bias but not the variance bias, isolating the latter and revealing that high- (respectively, low-) variance clusters are undersampled (resp., oversampled) relative to exact softmax ($R_i^{(N)}(q) \neq 1$,~$p < 0.01$). SD-2LS corrects these biases.}
\end{figure}

\begin{figure}[th!]
\centering
\begin{subfigure}[t]{0.32\linewidth}
    \centering
    \includegraphics[width=\linewidth]{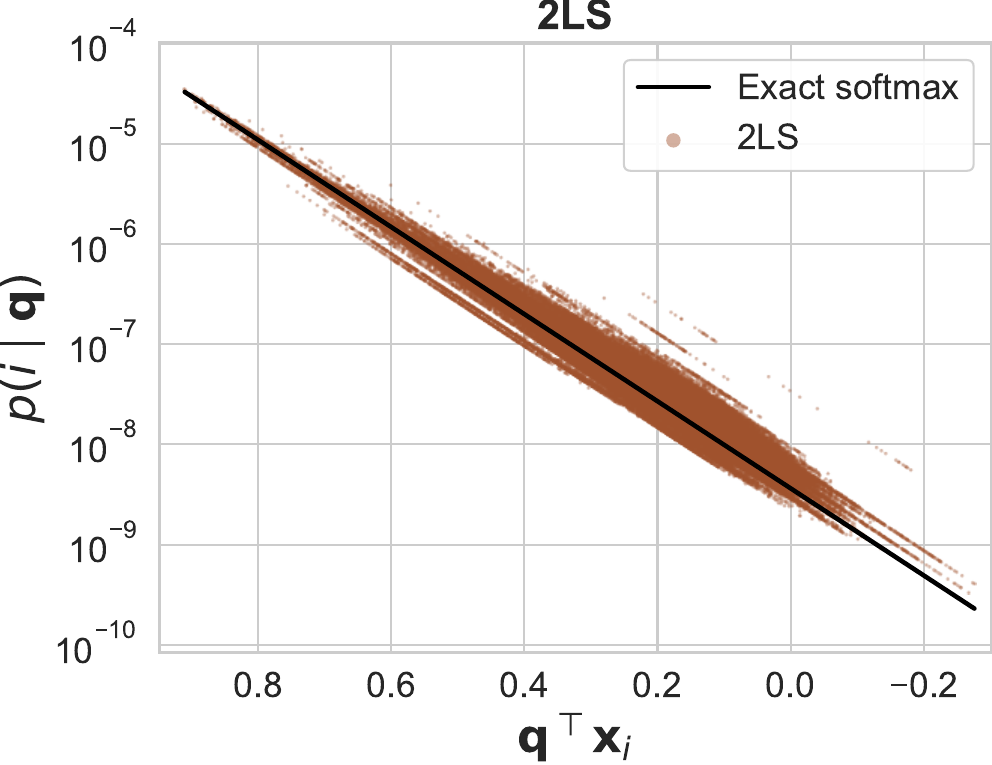}
\end{subfigure}
\hfill
\begin{subfigure}[t]{0.32\linewidth}
    \centering
    \includegraphics[width=\linewidth]{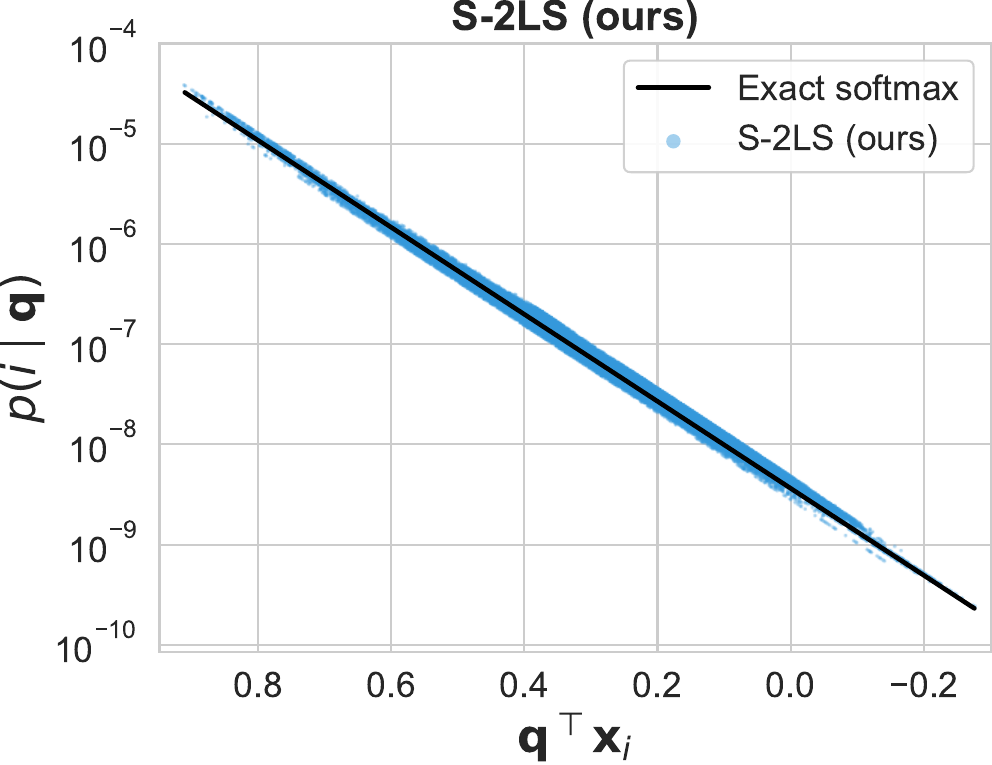}
\end{subfigure}
\hfill
\begin{subfigure}[t]{0.32\linewidth}
    \centering
    \includegraphics[width=\linewidth]{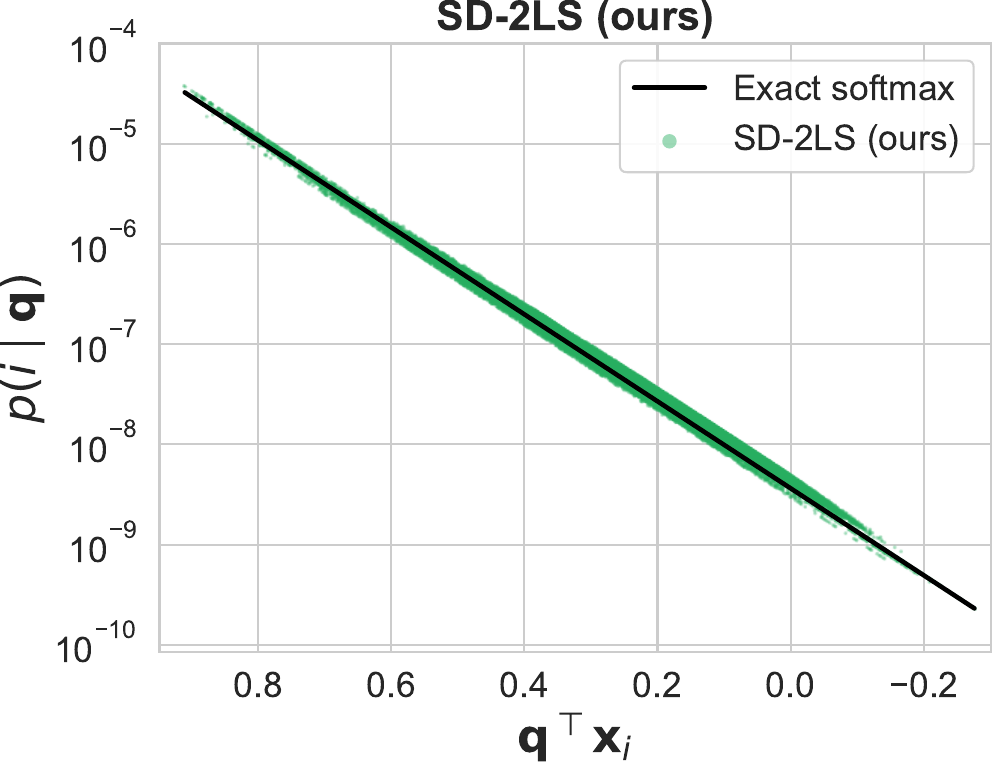}
\end{subfigure}
\caption{Sampling probability on \textbf{YAMBDA} ($\tau = 0.1$) per item as a function of query-item dot-product similarity for a fixed query. Standard 2LS assigns widely different probabilities to items with identical similarity, while S-2LS and SD-2LS markedly reduce this dispersion.}
\end{figure}

\clearpage


\begin{figure}[t]
\vspace{2.5cm}
\centering
\begin{subfigure}[t]{0.325\linewidth}
    \centering
    \includegraphics[width=\linewidth]{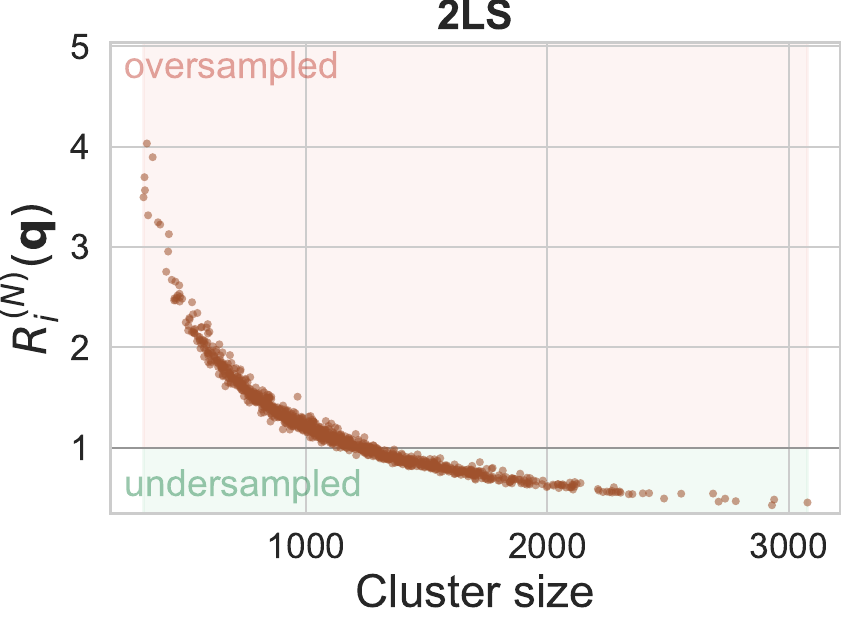}
\end{subfigure}
\hfill
\begin{subfigure}[t]{0.325\linewidth}
    \centering
    \includegraphics[width=\linewidth]{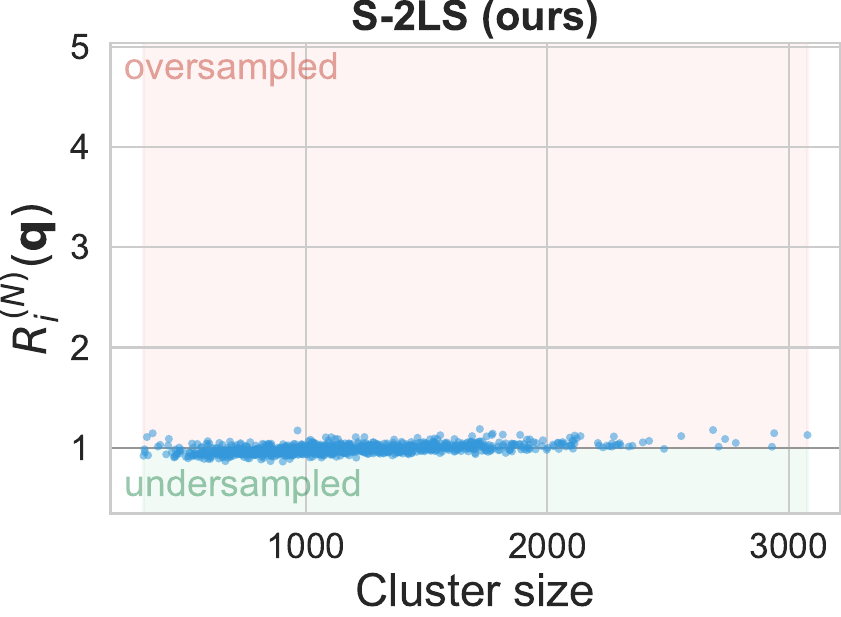}
\end{subfigure}
\hfill
\begin{subfigure}[t]{0.325\linewidth}
    \centering
    \includegraphics[width=\linewidth]{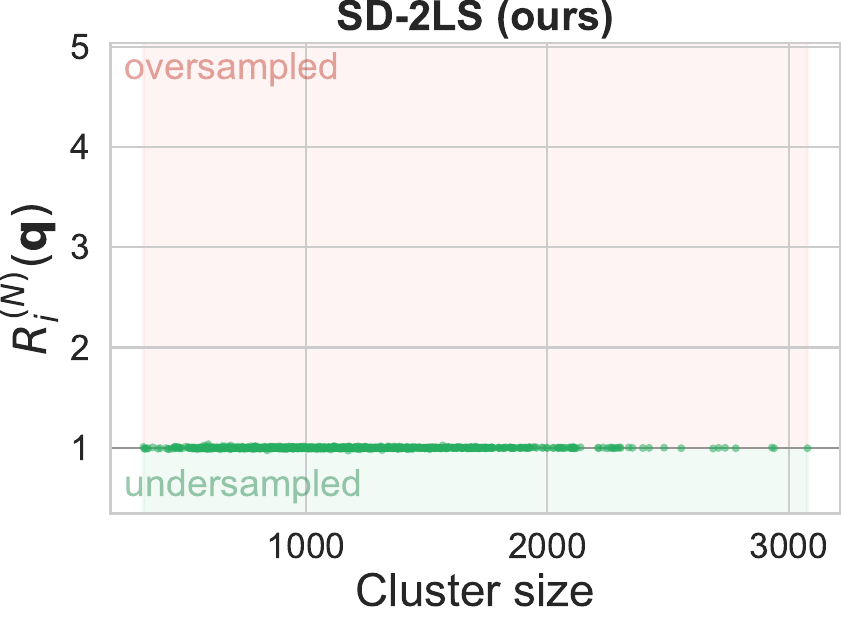}
\end{subfigure}
\caption{Sampling ratio on \textbf{GloVe-100} ($\tau = 0.1$) as a function of cluster size, averaged over queries. Standard 2LS  undersamples large clusters and oversamples small ones relative to exact softmax ($R_i^{(N)}(q) \neq 1$, p-value $p < 0.01$), while S-2LS and SD-2LS correct these biases.}
\end{figure}

\begin{figure}[t]
\centering
\begin{subfigure}[t]{0.32\linewidth}
    \centering
    \includegraphics[width=\linewidth]{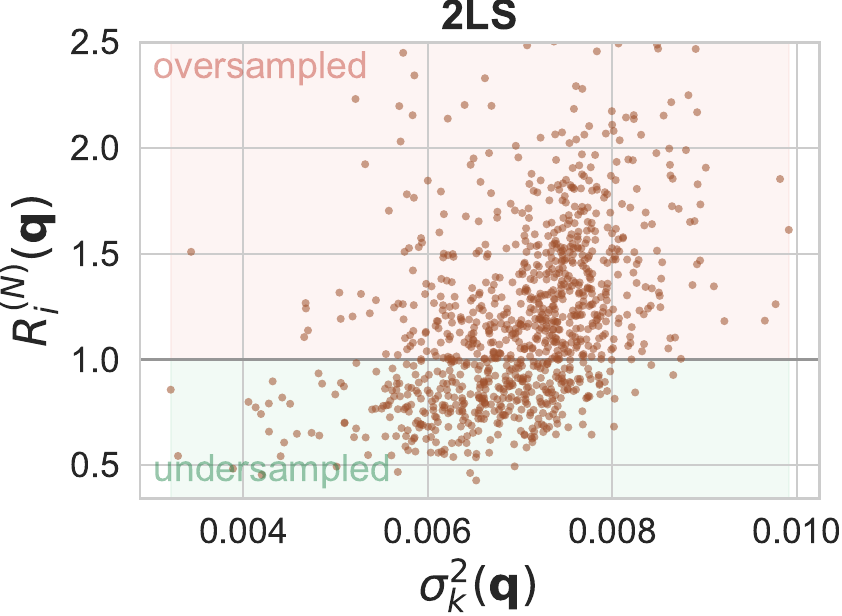}
\end{subfigure}
\hfill
\begin{subfigure}[t]{0.32\linewidth}
    \centering
    \includegraphics[width=\linewidth]{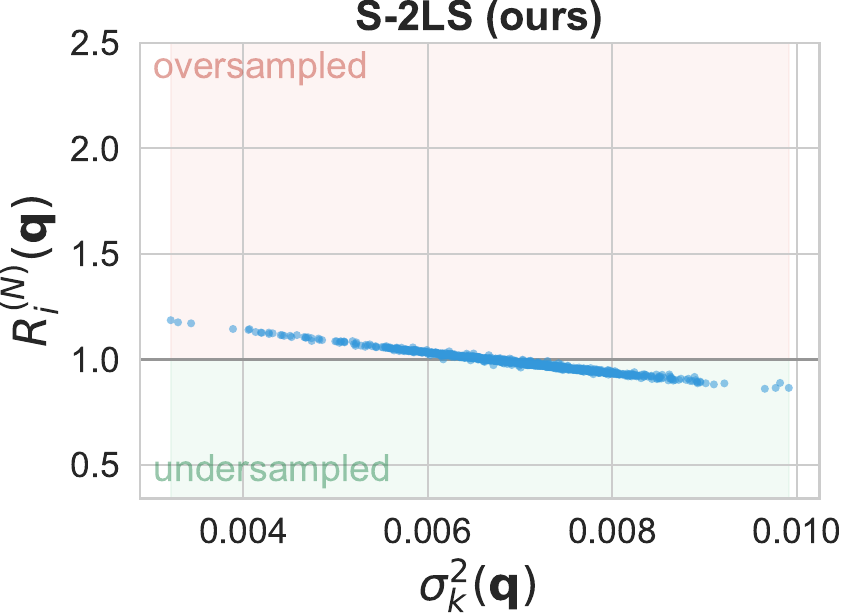}
\end{subfigure}
\hfill
\begin{subfigure}[t]{0.32\linewidth}
    \centering
    \includegraphics[width=\linewidth]{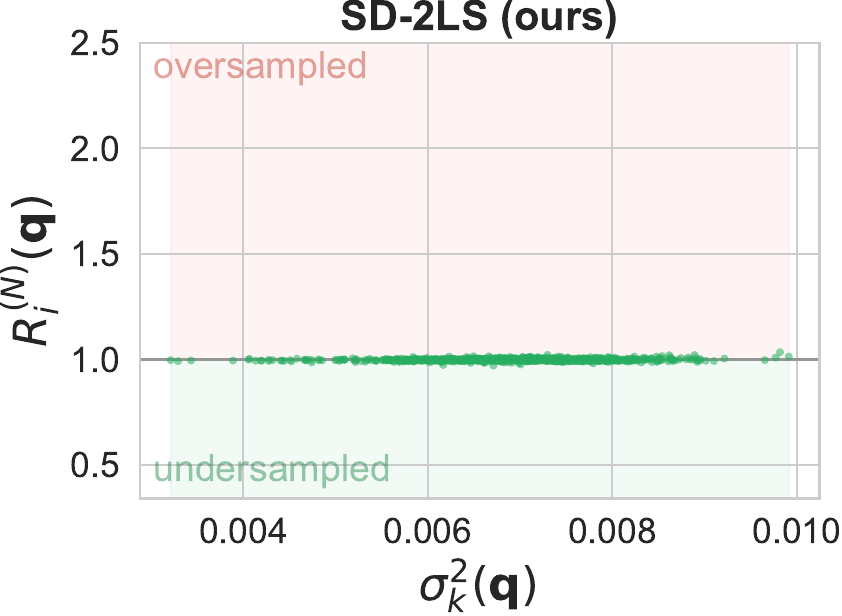}
\end{subfigure}
\caption{Sampling ratio on \textbf{GloVe-100} ($\tau = 0.1$) as a function of intra-cluster similarity variance, averaged over queries. The effect is entangled with cluster-size bias for 2LS; S-2LS corrects the cluster-size bias but not the variance bias, isolating the latter and revealing that high- (respectively, low-) variance clusters are undersampled (resp., oversampled) relative to exact softmax ($R_i^{(N)}(q) \neq 1$,~$p < 0.01$). SD-2LS corrects these biases.}
\end{figure}

\begin{figure}[th!]
\centering
\begin{subfigure}[t]{0.32\linewidth}
    \centering
    \includegraphics[width=\linewidth]{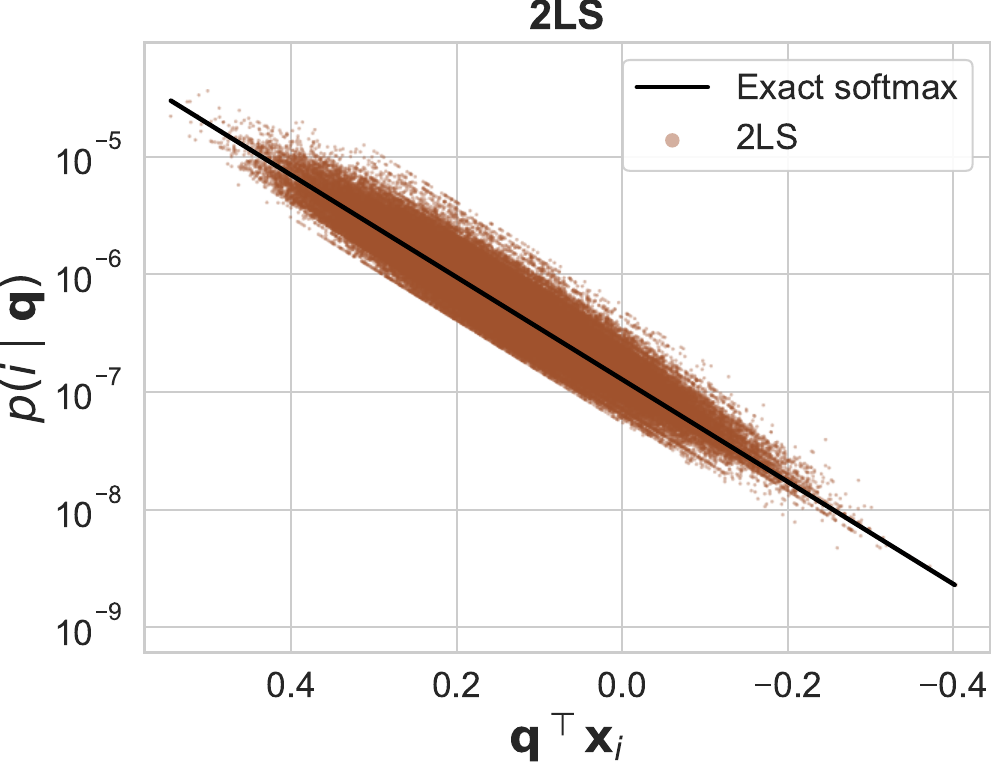}
\end{subfigure}
\hfill
\begin{subfigure}[t]{0.32\linewidth}
    \centering
    \includegraphics[width=\linewidth]{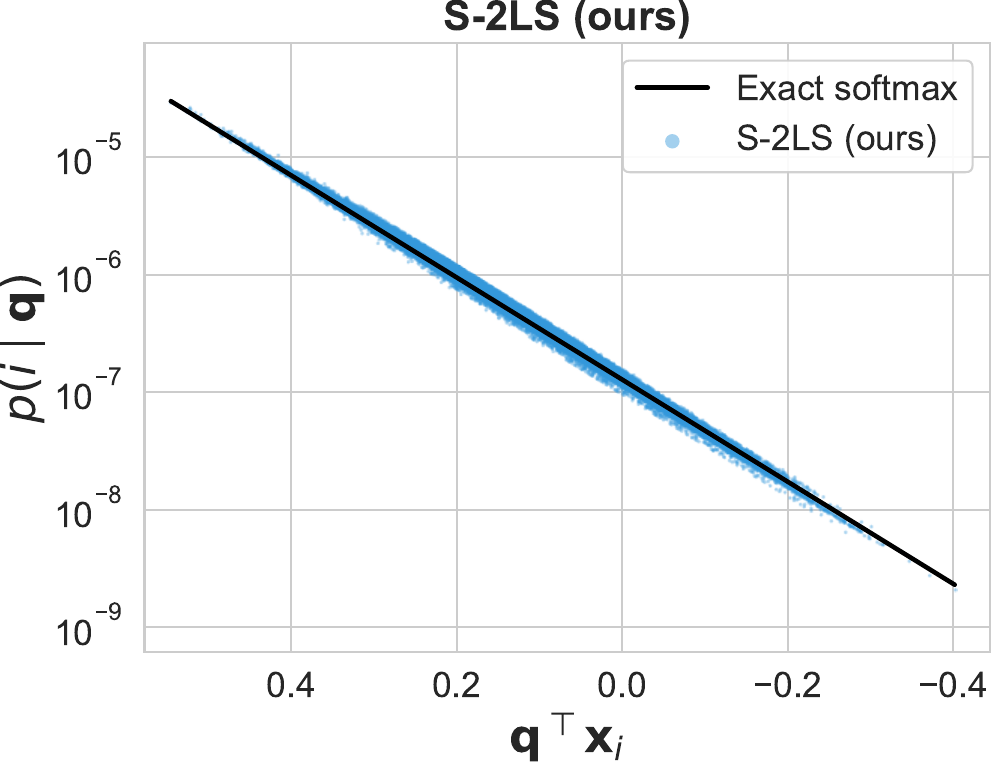}
\end{subfigure}
\hfill
\begin{subfigure}[t]{0.32\linewidth}
    \centering
    \includegraphics[width=\linewidth]{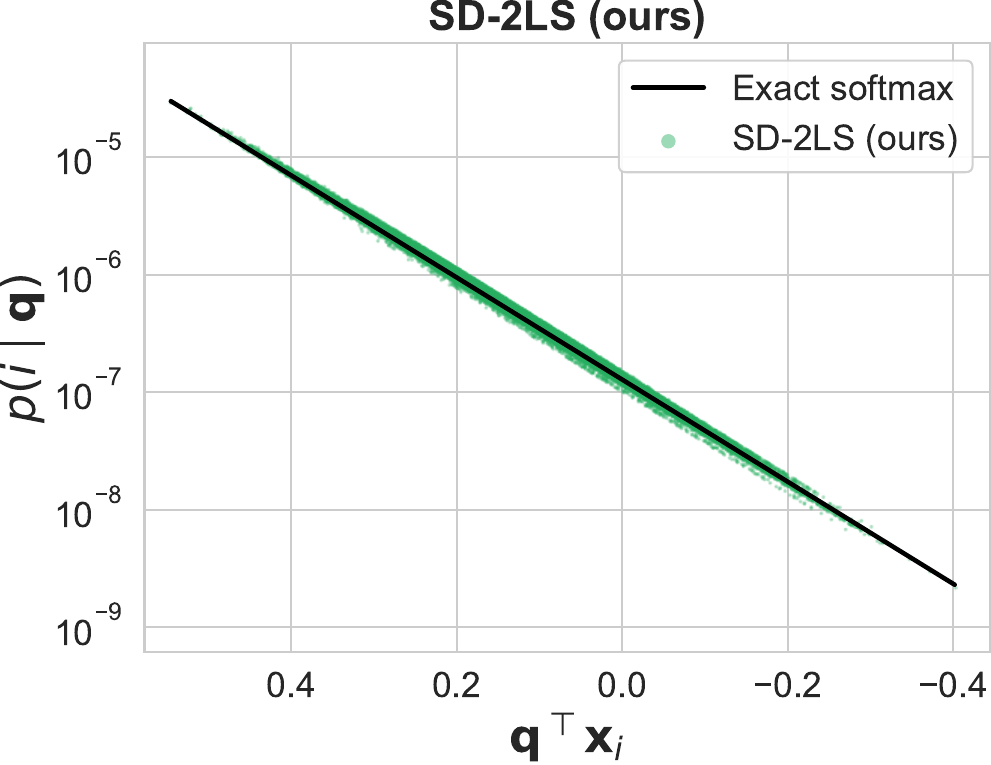}
\end{subfigure}
\caption{Sampling probability on \textbf{GloVe-100} ($\tau = 0.1$) per item as a function of query-item dot-product similarity for a fixed query. Standard 2LS assigns widely different probabilities to items with identical similarity, while S-2LS and SD-2LS markedly reduce this dispersion.}
\end{figure}

\clearpage


\begin{figure}[t]
\vspace{2.5cm}
\centering
\begin{subfigure}[t]{0.325\linewidth}
    \centering
    \includegraphics[width=\linewidth]{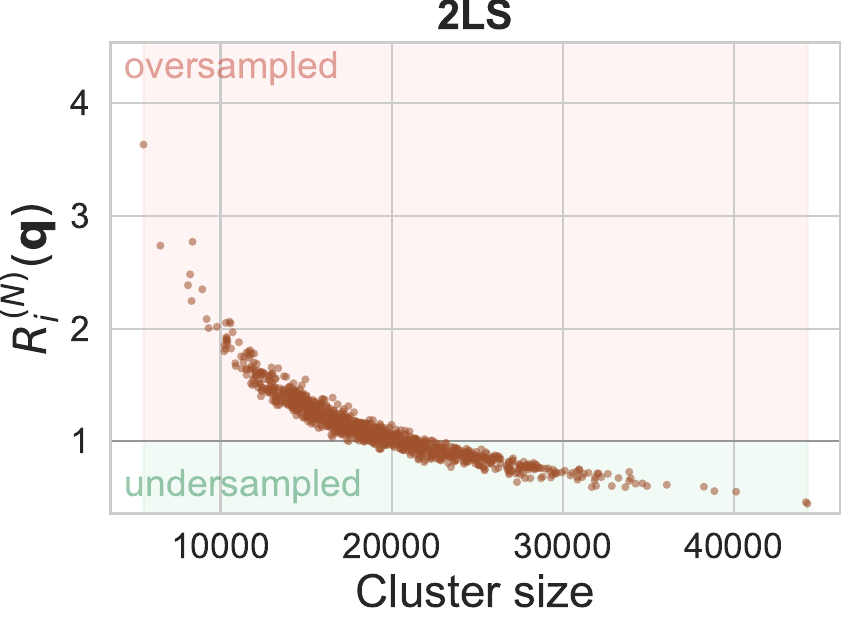}
\end{subfigure}
\hfill
\begin{subfigure}[t]{0.325\linewidth}
    \centering
    \includegraphics[width=\linewidth]{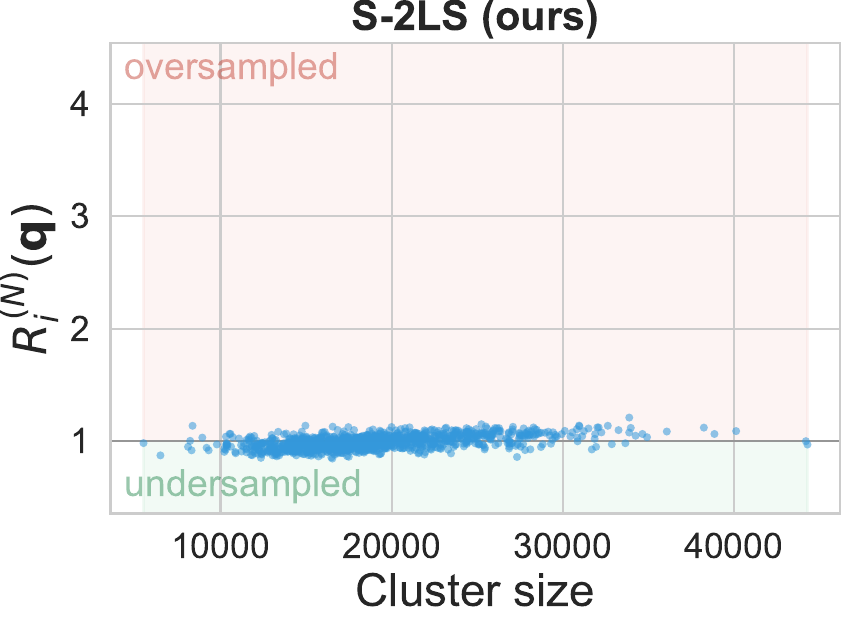}
\end{subfigure}
\hfill
\begin{subfigure}[t]{0.325\linewidth}
    \centering
    \includegraphics[width=\linewidth]{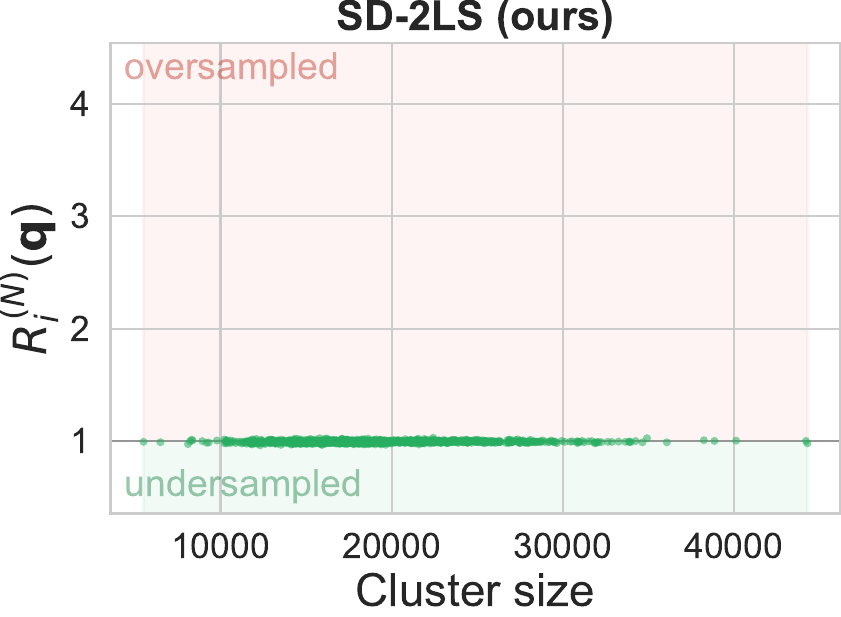}
\end{subfigure}
\caption{Sampling ratio on \textbf{VK-LSVD} ($\tau = 0.1$) as a function of cluster size, averaged over queries. Standard 2LS  undersamples large clusters and oversamples small ones relative to exact softmax ($R_i^{(N)}(q) \neq 1$, p-value $p < 0.01$), while S-2LS and SD-2LS correct these biases.}
\end{figure}

\begin{figure}[t]
\centering
\begin{subfigure}[t]{0.32\linewidth}
    \centering
    \includegraphics[width=\linewidth]{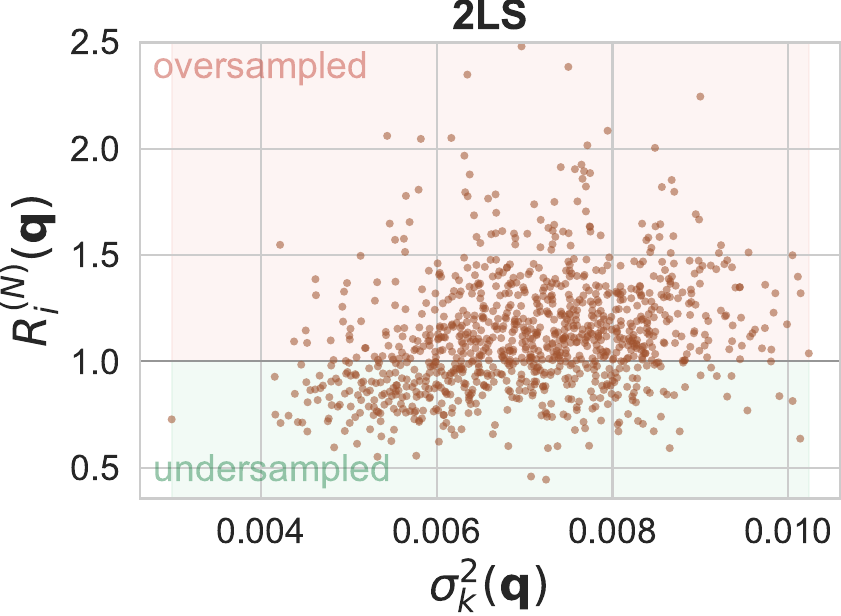}
\end{subfigure}
\hfill
\begin{subfigure}[t]{0.32\linewidth}
    \centering
    \includegraphics[width=\linewidth]{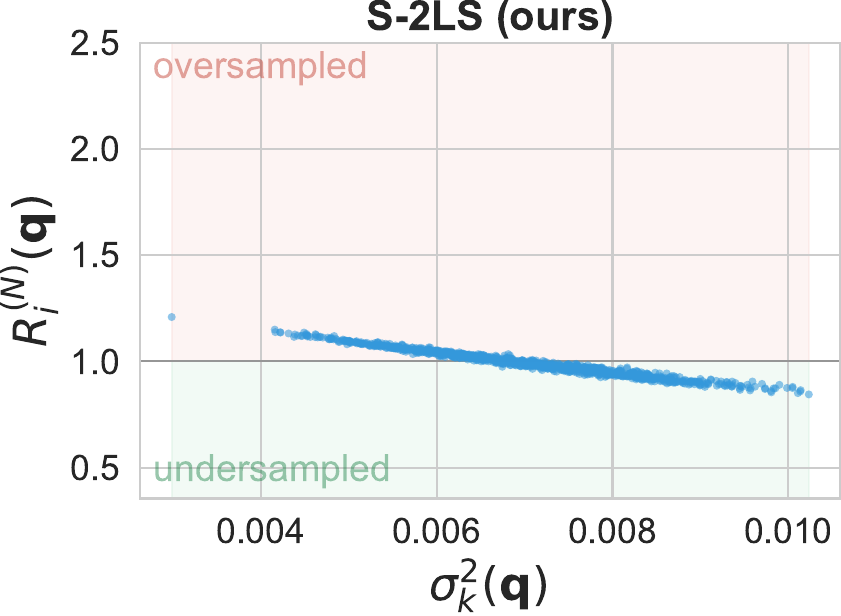}
\end{subfigure}
\hfill
\begin{subfigure}[t]{0.32\linewidth}
    \centering
    \includegraphics[width=\linewidth]{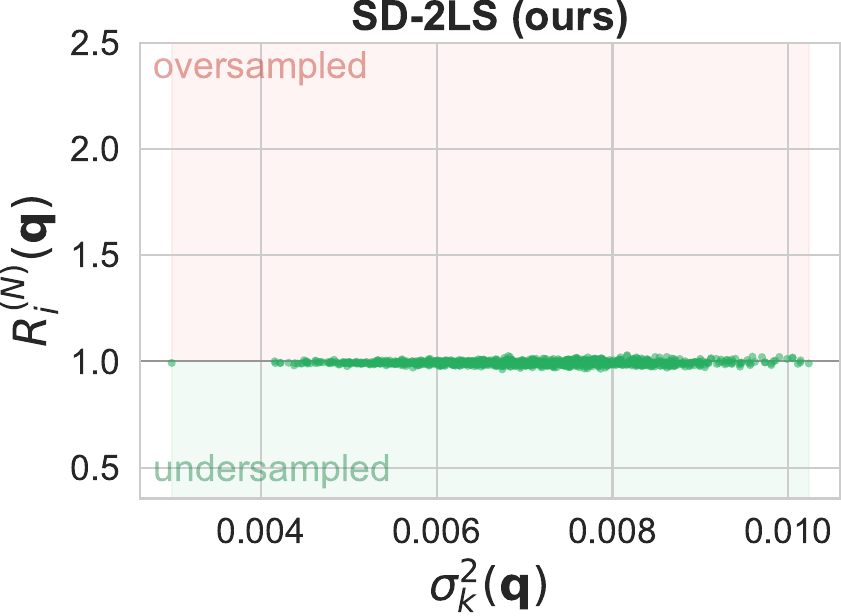}
\end{subfigure}
\caption{Sampling ratio on \textbf{VK-LSVD} ($\tau = 0.1$) as a function of intra-cluster similarity variance, averaged over queries. The effect is entangled with cluster-size bias for 2LS; S-2LS corrects the cluster-size bias but not the variance bias, isolating the latter and revealing that high- (respectively, low-) variance clusters are undersampled (resp., oversampled) relative to exact softmax ($R_i^{(N)}(q) \neq 1$,~$p < 0.01$). SD-2LS corrects these biases.}
\end{figure}

\begin{figure}[th!]
\centering
\begin{subfigure}[t]{0.32\linewidth}
    \centering
    \includegraphics[width=\linewidth]{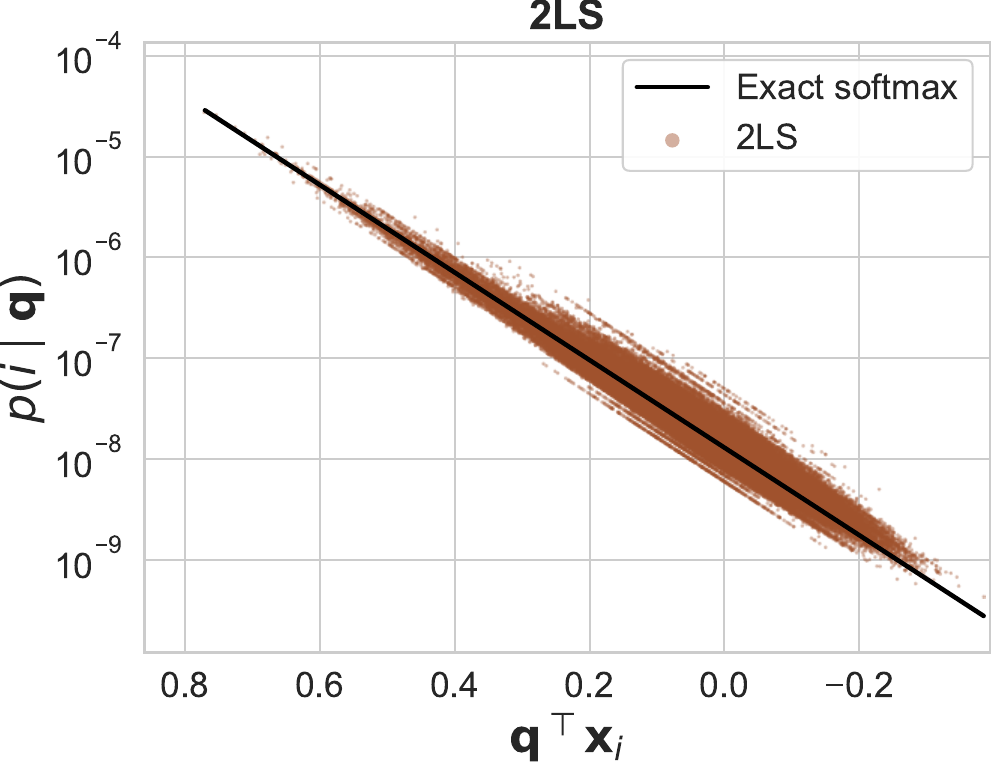}
\end{subfigure}
\hfill
\begin{subfigure}[t]{0.32\linewidth}
    \centering
    \includegraphics[width=\linewidth]{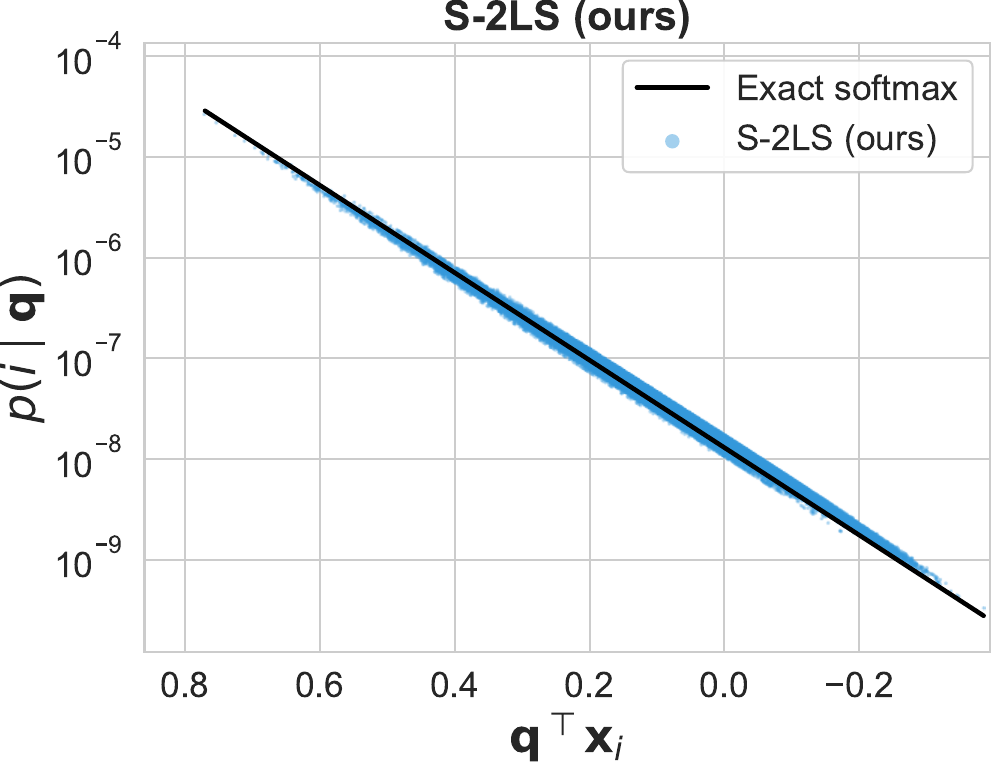}
\end{subfigure}
\hfill
\begin{subfigure}[t]{0.32\linewidth}
    \centering
    \includegraphics[width=\linewidth]{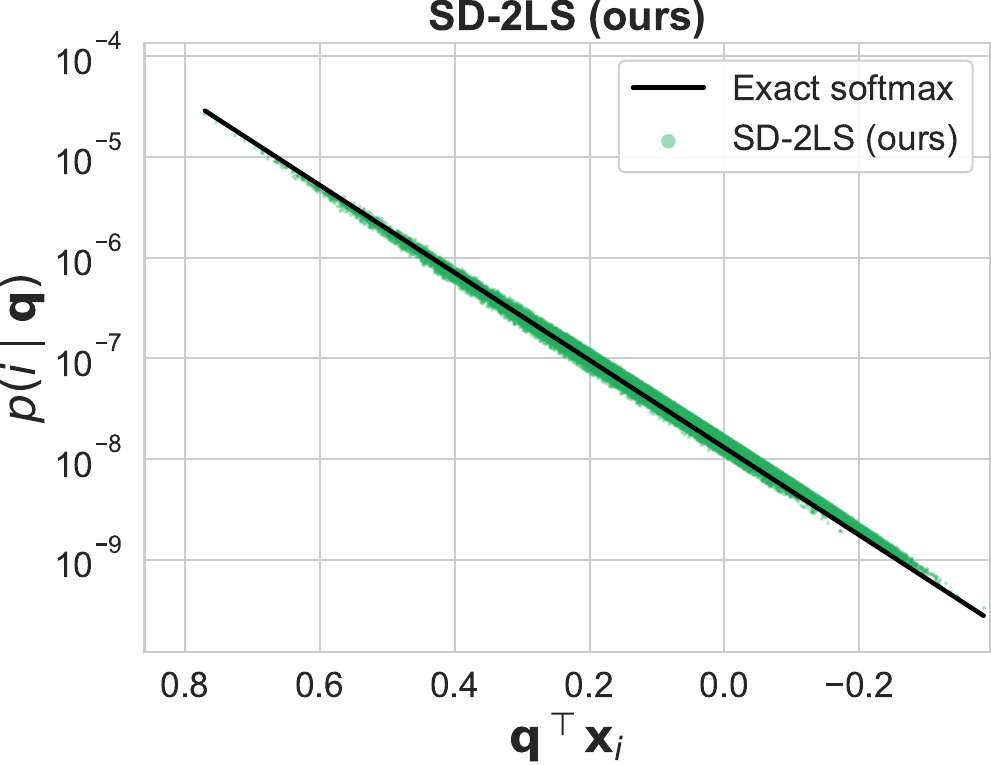}
\end{subfigure}
\caption{Sampling probability on \textbf{VK-LSVD} ($\tau = 0.1$) per item as a function of query-item dot-product similarity for a fixed query. Standard 2LS assigns widely different probabilities to items with identical similarity, while S-2LS and SD-2LS markedly reduce this dispersion.}
\end{figure}

\clearpage


\begin{figure}[t]
\vspace{2.5cm}
\centering
\begin{subfigure}[t]{0.325\linewidth}
    \centering
    \includegraphics[width=\linewidth]{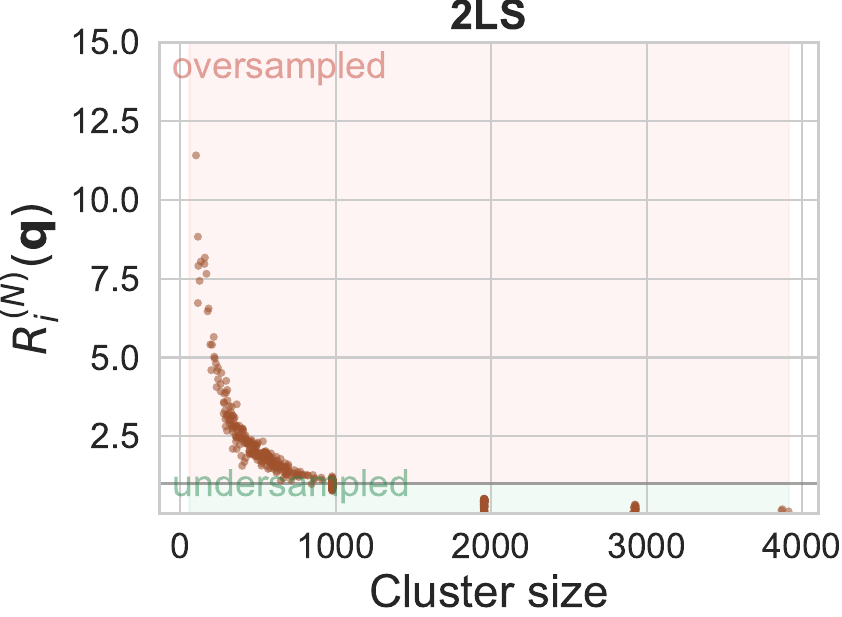}
\end{subfigure}
\hfill
\begin{subfigure}[t]{0.325\linewidth}
    \centering
    \includegraphics[width=\linewidth]{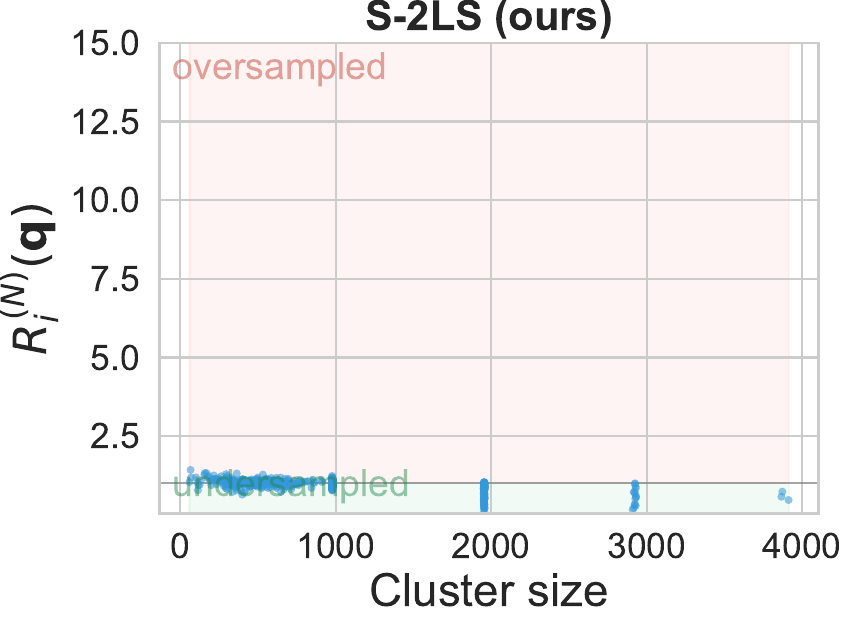}
\end{subfigure}
\hfill
\begin{subfigure}[t]{0.325\linewidth}
    \centering
    \includegraphics[width=\linewidth]{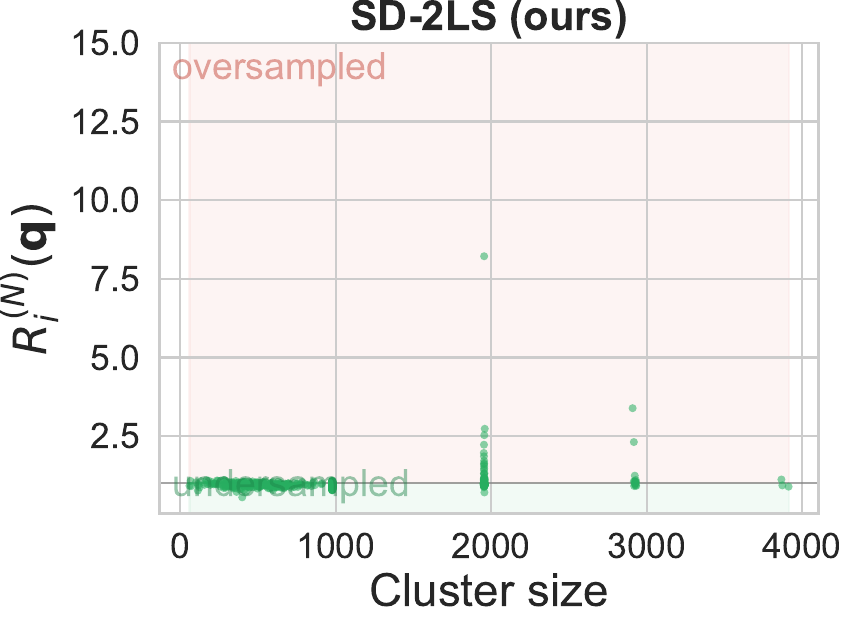}
\end{subfigure}
\caption{Sampling ratio on \textbf{Synth-Balanced} ($\tau = 0.1$) as a function of cluster size, averaged over queries. Standard 2LS  undersamples large clusters and oversamples small ones relative to exact softmax ($R_i^{(N)}(q) \neq 1$, p-value $p < 0.01$), while S-2LS and SD-2LS correct these biases.}
\end{figure}

\begin{figure}[t]
\centering
\begin{subfigure}[t]{0.32\linewidth}
    \centering
    \includegraphics[width=\linewidth]{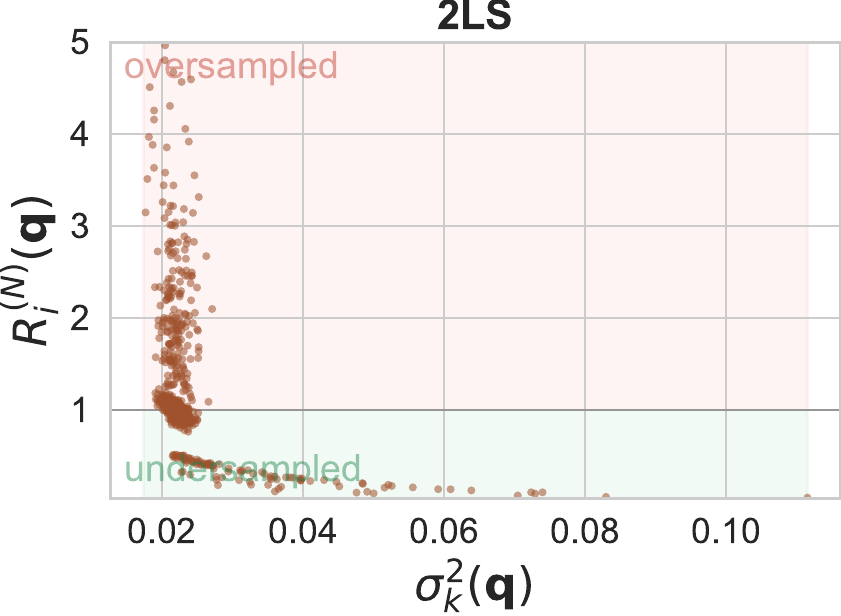}
\end{subfigure}
\hfill
\begin{subfigure}[t]{0.32\linewidth}
    \centering
    \includegraphics[width=\linewidth]{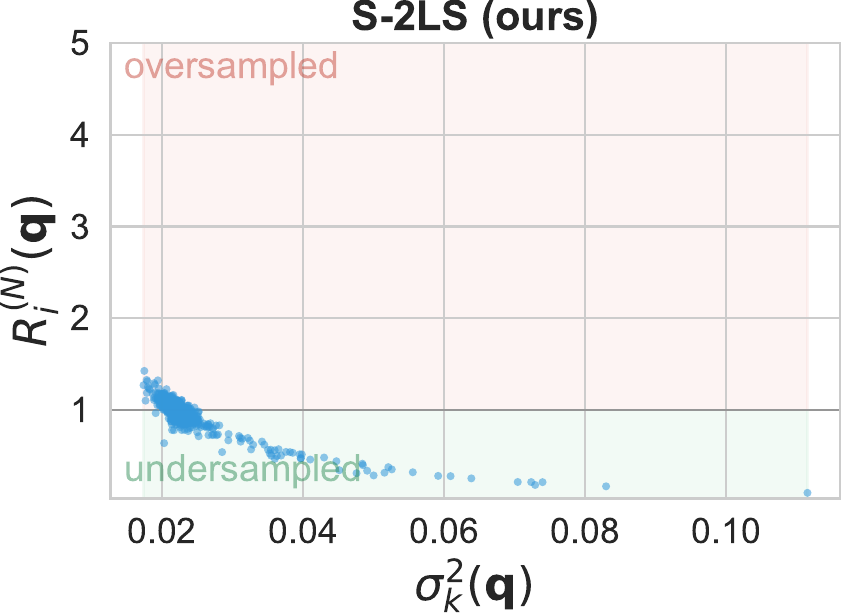}
\end{subfigure}
\hfill
\begin{subfigure}[t]{0.32\linewidth}
    \centering
    \includegraphics[width=\linewidth]{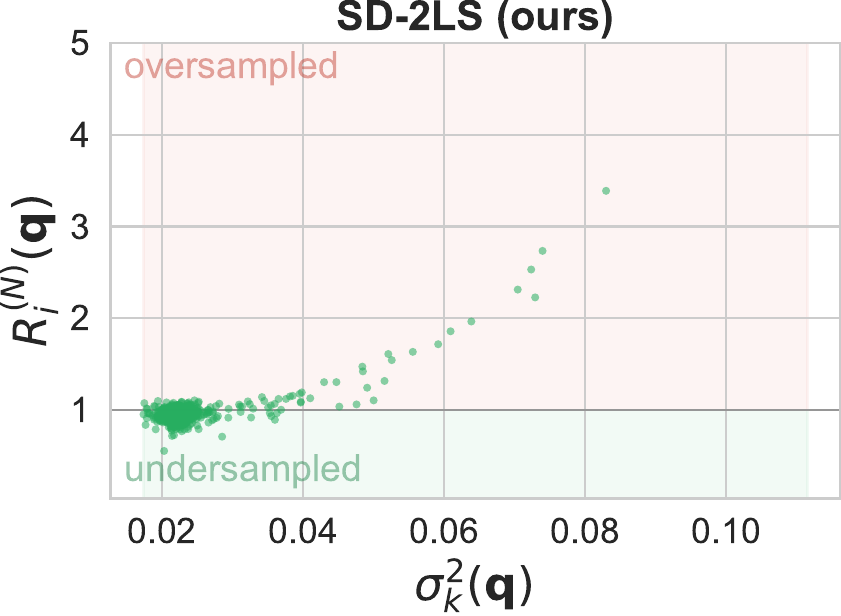}
\end{subfigure}
\caption{Sampling ratio on \textbf{Synth-Balanced} ($\tau = 0.1$) as a function of intra-cluster similarity variance, averaged over queries. The effect is entangled with cluster-size bias for 2LS; S-2LS corrects the cluster-size bias but not the variance bias, isolating the latter and revealing that high- (respectively, low-) variance clusters are undersampled (resp., oversampled) relative to exact softmax ($R_i^{(N)}(q) \neq 1$,~$p < 0.01$). SD-2LS corrects these biases for the vast majority of points.}
\end{figure}

\begin{figure}[th!]
\centering
\begin{subfigure}[t]{0.32\linewidth}
    \centering
    \includegraphics[width=\linewidth]{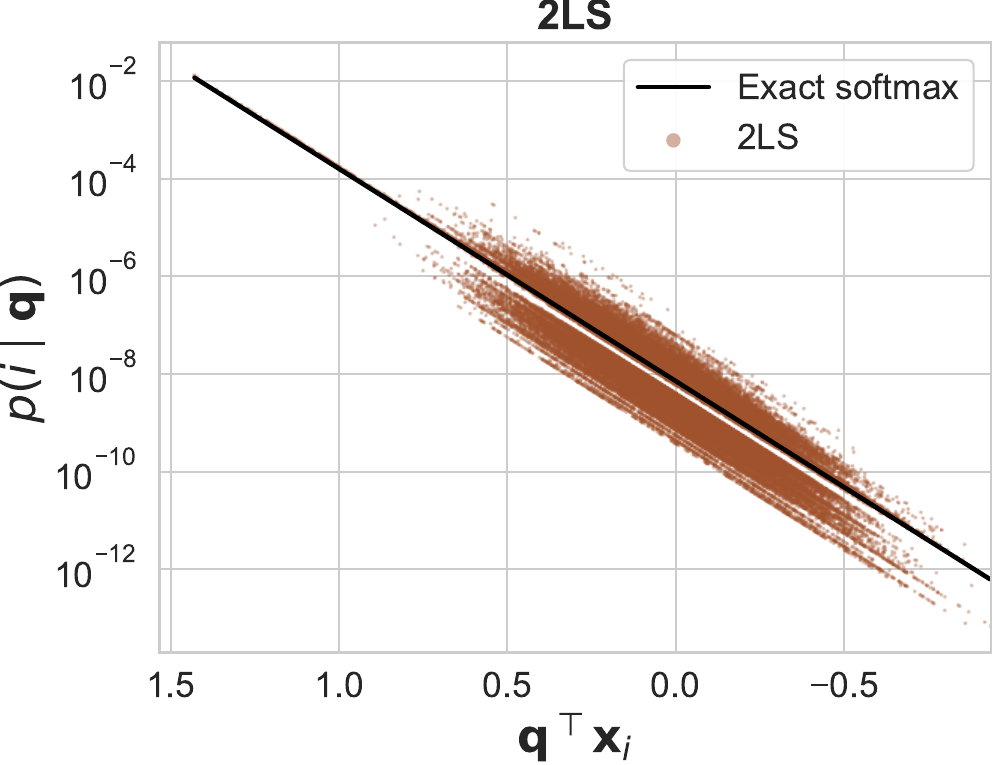}
\end{subfigure}
\hfill
\begin{subfigure}[t]{0.32\linewidth}
    \centering
    \includegraphics[width=\linewidth]{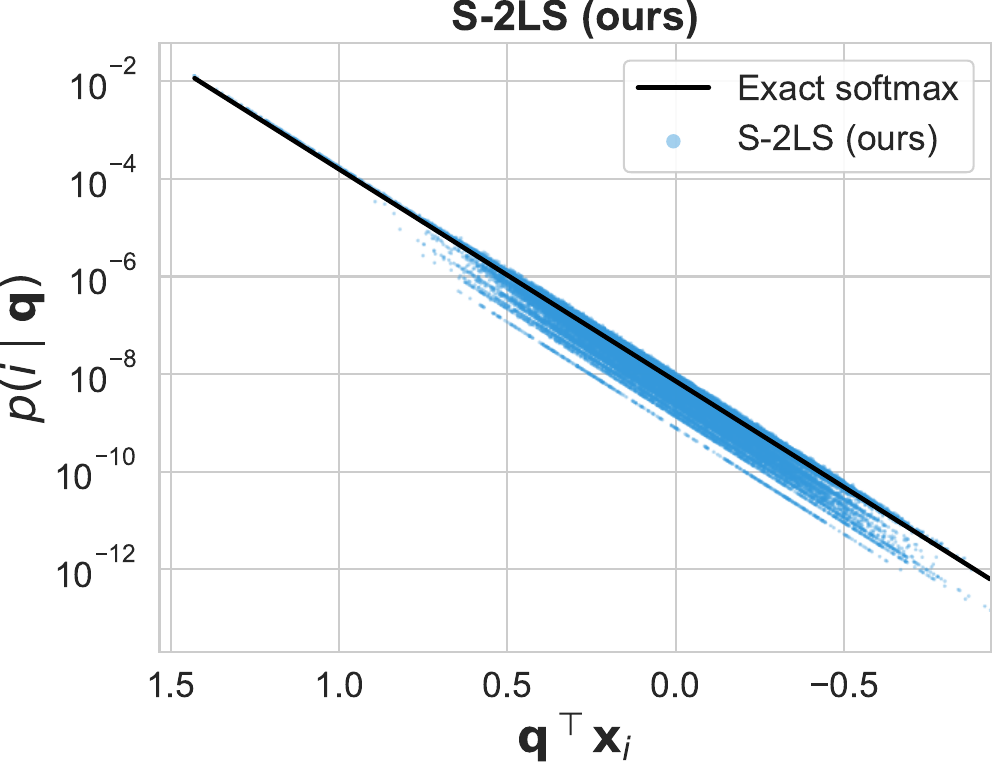}
\end{subfigure}
\hfill
\begin{subfigure}[t]{0.32\linewidth}
    \centering
    \includegraphics[width=\linewidth]{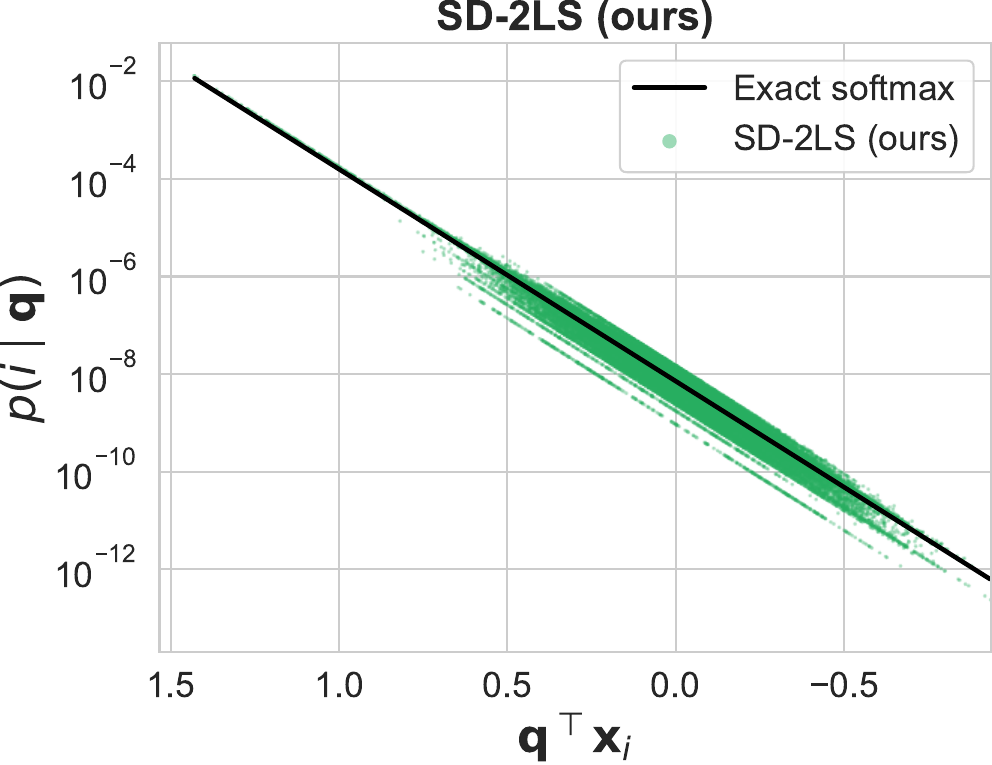}
\end{subfigure}
\caption{Sampling probability on \textbf{Synth-Balanced} ($\tau = 0.1$) per item as a function of query-item dot-product similarity for a fixed query. Standard 2LS assigns widely different probabilities to items with identical similarity, while S-2LS and SD-2LS markedly reduce this dispersion.}
\end{figure}

\clearpage


\begin{figure}[t]
\vspace{2.5cm}
\centering
\begin{subfigure}[t]{0.325\linewidth}
    \centering
    \includegraphics[width=\linewidth]{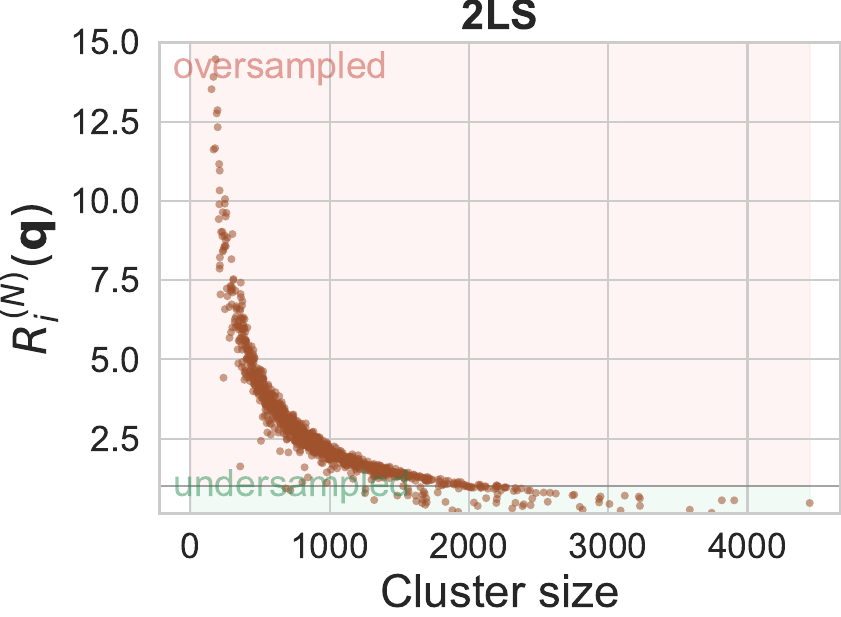}
\end{subfigure}
\hfill
\begin{subfigure}[t]{0.325\linewidth}
    \centering
    \includegraphics[width=\linewidth]{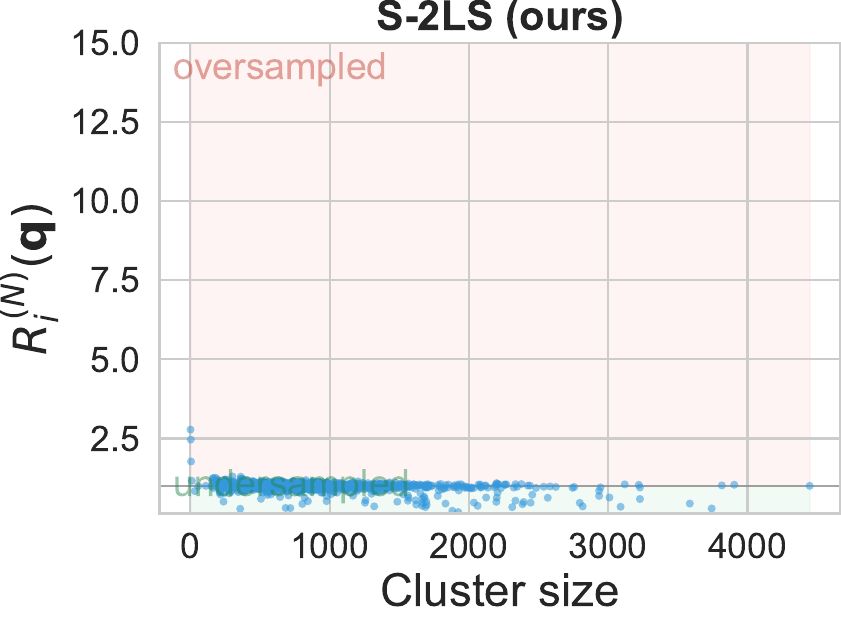}
\end{subfigure}
\hfill
\begin{subfigure}[t]{0.325\linewidth}
    \centering
    \includegraphics[width=\linewidth]{figures_pdf/synth-balanced/iw_size_sd2ls.pdf}
\end{subfigure}
\caption{Sampling ratio on \textbf{Synth-Unbalanced} ($\tau = 0.1$) as a function of cluster size, averaged over queries. Standard 2LS  undersamples large clusters and oversamples small ones relative to exact softmax ($R_i^{(N)}(q) \neq 1$, p-value $p < 0.01$), while S-2LS and SD-2LS correct these biases.}
\end{figure}

\begin{figure}[t]
\centering
\begin{subfigure}[t]{0.32\linewidth}
    \centering
    \includegraphics[width=\linewidth]{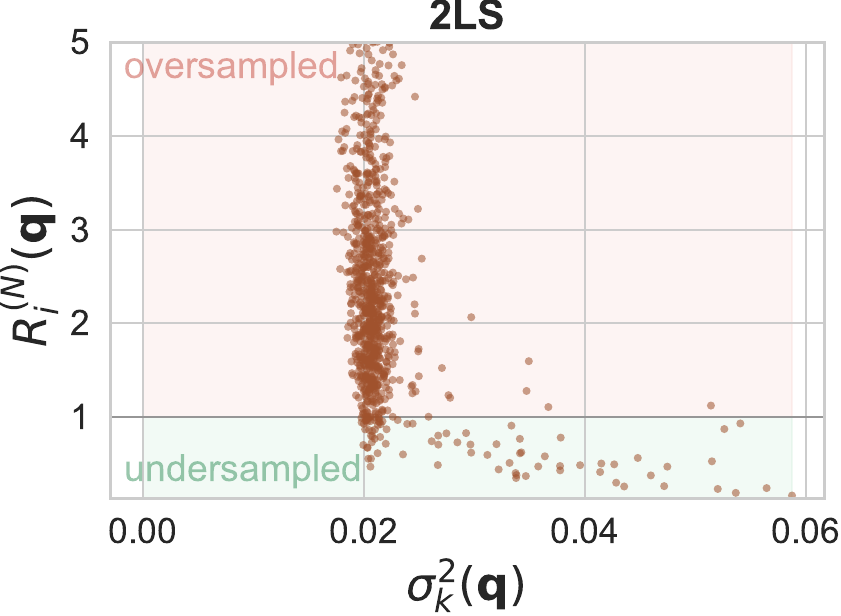}
\end{subfigure}
\hfill
\begin{subfigure}[t]{0.32\linewidth}
    \centering
    \includegraphics[width=\linewidth]{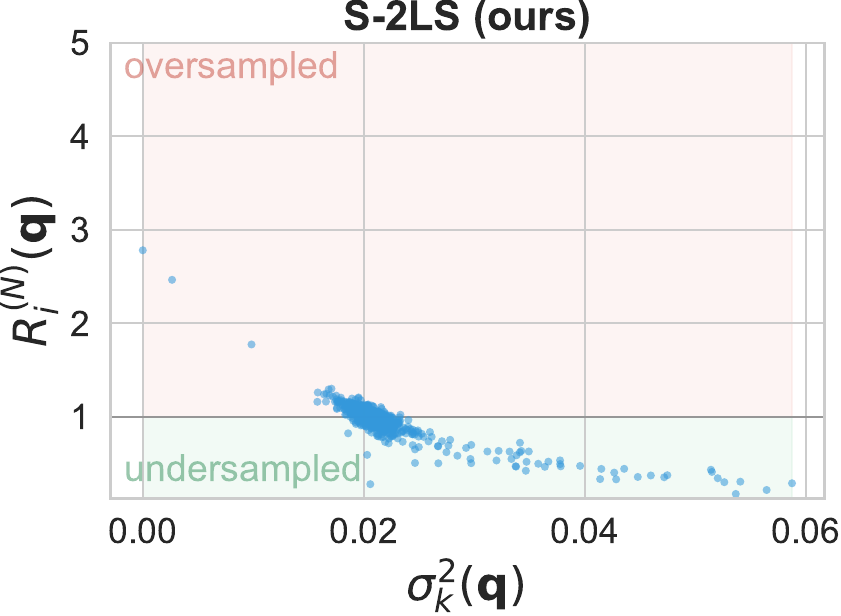}
\end{subfigure}
\hfill
\begin{subfigure}[t]{0.32\linewidth}
    \centering
    \includegraphics[width=\linewidth]{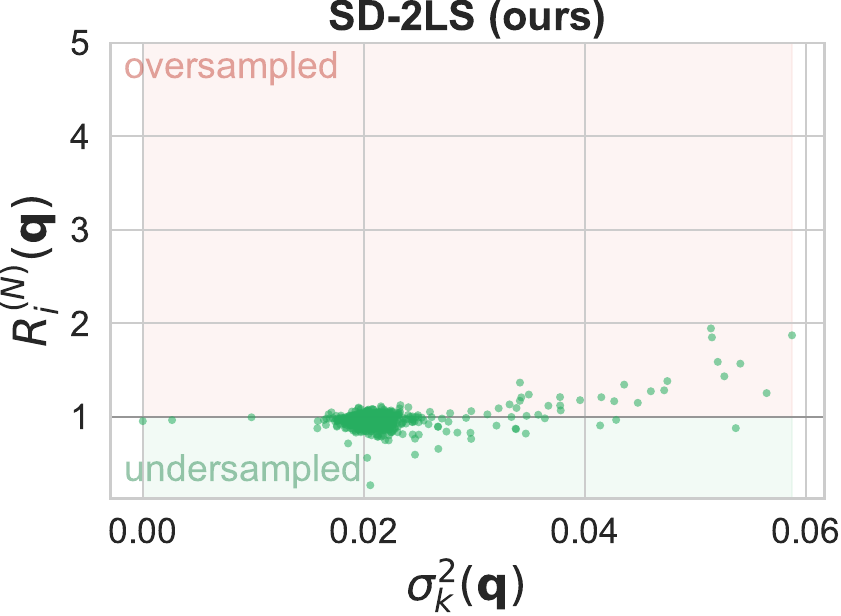}
\end{subfigure}
\caption{Sampling ratio on \textbf{Synth-Unbalanced} ($\tau = 0.1$) as a function of intra-cluster similarity variance, averaged over queries.  The effect is entangled with cluster-size bias for 2LS; S-2LS corrects the cluster-size bias but not the variance bias, isolating the latter and revealing that high- (respectively, low-) variance clusters are undersampled (resp., oversampled) relative to exact softmax ($R_i^{(N)}(q) \neq 1$,~$p < 0.01$). SD-2LS corrects these biases for the vast majority of points.}
\end{figure}

\begin{figure}[th!]
\centering
\begin{subfigure}[t]{0.32\linewidth}
    \centering
    \includegraphics[width=\linewidth]{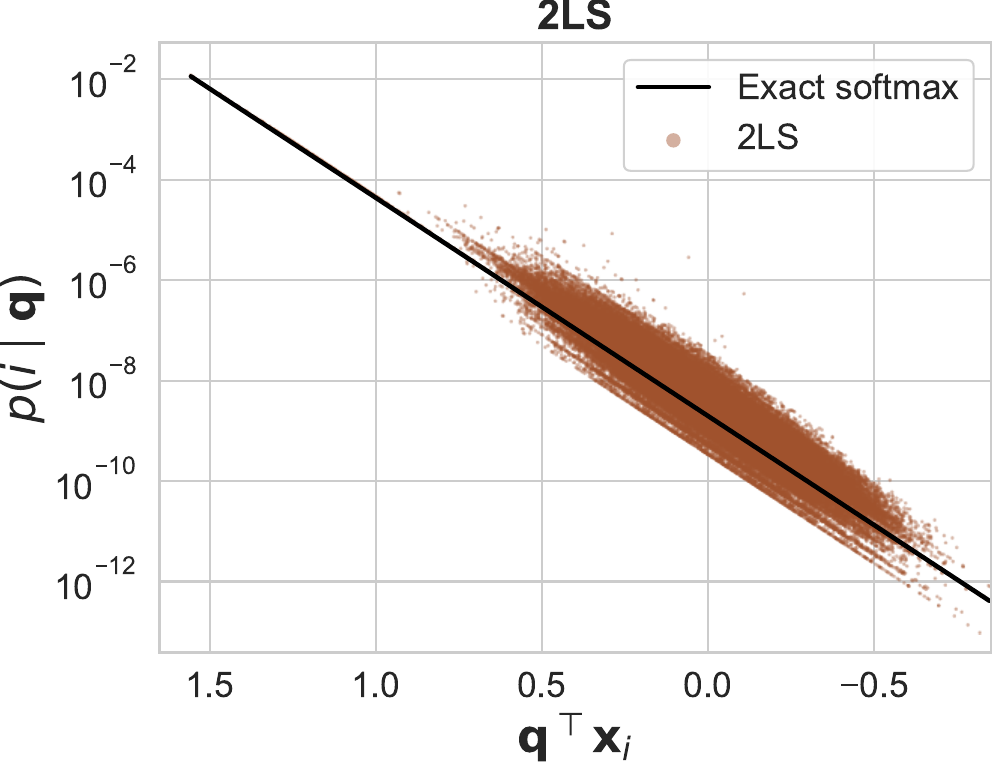}
\end{subfigure}
\hfill
\begin{subfigure}[t]{0.32\linewidth}
    \centering
    \includegraphics[width=\linewidth]{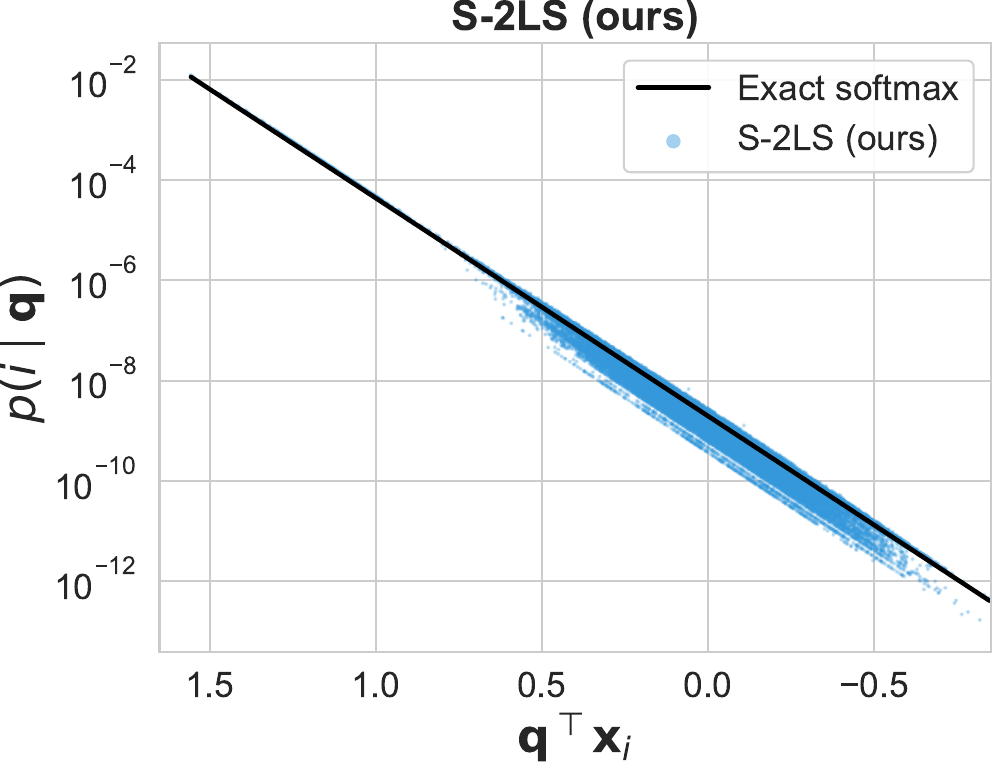}
\end{subfigure}
\hfill
\begin{subfigure}[t]{0.32\linewidth}
    \centering
    \includegraphics[width=\linewidth]{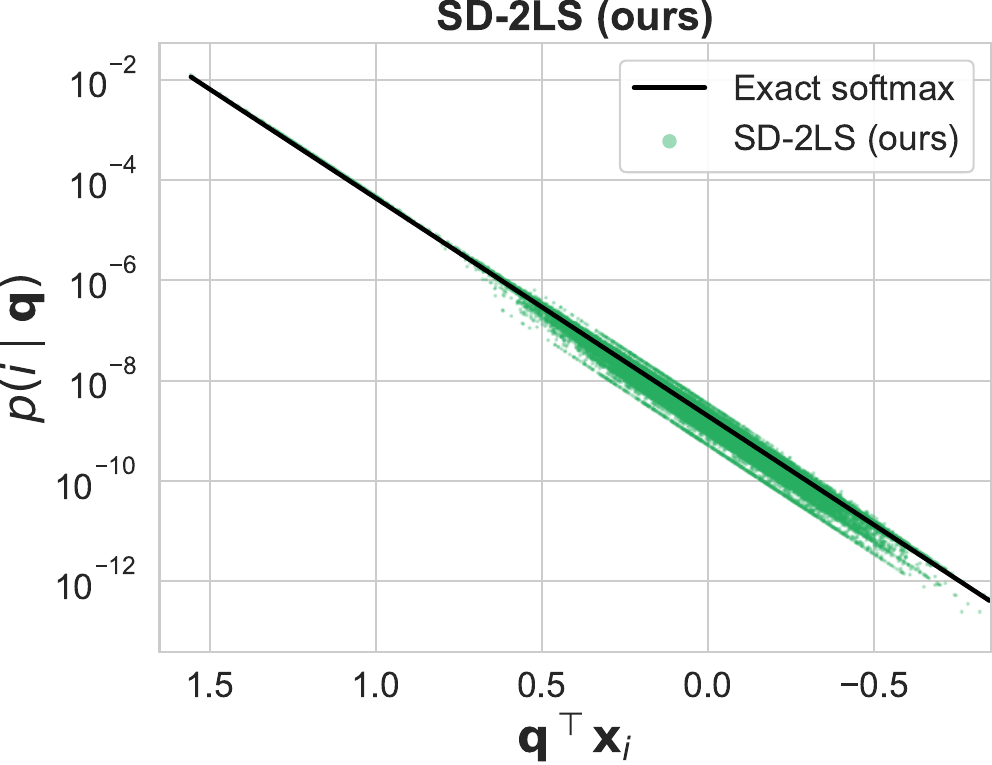}
\end{subfigure}
\caption{Sampling probability on \textbf{Synth-Unbalanced} ($\tau = 0.1$) per item as a function of query-item dot-product similarity for a fixed query. Standard 2LS assigns widely different probabilities to items with identical similarity, while S-2LS and SD-2LS markedly reduce this dispersion.}
\end{figure}


\end{document}